\documentclass[10pt,twocolumn,letterpaper]{article}

\usepackage[pagenumbers]{wacv}              %

\usepackage{graphicx}   %
\usepackage{multirow}
\usepackage{array}
\usepackage{adjustbox}
\usepackage{booktabs}   %
\usepackage{amsmath}    %
\usepackage{amssymb}    %
\usepackage{xcolor}   %

\definecolor{wacvblue}{rgb}{0.21,0.49,0.74}
\usepackage[pagebackref,breaklinks,colorlinks,allcolors=wacvblue]{hyperref}

\def\wacvPaperID{2868} %
\def\confName{WACV}
\def\confYear{2027}

\title{Do We Still Need Rotation-Invariant Architectures? A Statistical Robustness Benchmark for Point Cloud Classification}

\author{Abdulla Nasir Chowdhury \quad Gulpi Qorik Oktagalu Pratamasunu\\
University of Tsukuba\\
{\tt\small \{abdullah.nasir,gulpi\}@cvlab.cs.tsukuba.ac.jp}
\and
Kazuhiro Fukui\\
University of Tsukuba\\
{\tt\small kfukui@cs.tsukuba.ac.jp}}

\AddToShipoutPicture*{%
  \AtPageUpperLeft{%
    \put(\LenToUnit{0.5\paperwidth},\LenToUnit{-0.45in}){%
      \makebox[0pt][c]{\parbox{0.86\paperwidth}{\centering\scriptsize
        Accepted to the 2027 IEEE/CVF Winter Conference on Applications of
        Computer Vision (WACV).\\
        \copyright~2027 IEEE. Personal use of this material is permitted.
        Permission from IEEE must be obtained for all other uses, in any
        current or future media, including reprinting/republishing this
        material for advertising or promotional purposes, creating new
        collective works, for resale or redistribution to servers or lists,
        or reuse of any copyrighted component of this work in other works.}}}}}

\begin{document}
\maketitle
\begin{abstract}
A substantial line of work builds rotation invariance directly into
point cloud architectures. We benchmark 10 architectures (PointNet,
PointNet++, DGCNN, RIConv, RIConv++, RISurConv, PointMLP,
PointTransformer, and the equivariant VN-PointNet and VN-DGCNN)
across 5 domains under pose, noise, and occlusion,
and analyze the results with confirmatory hypothesis tests rather
than point estimates. The three rotation-invariant (RI) architectures
achieve the best pose robustness but rank among the lowest on noise robustness in 4 of 5 domains. We ask whether
this cost is intrinsic to invariance or specific to how invariance is
obtained. We retrain all 10 architectures unchanged, with a simple
invariance loss on
the embedding and three training seeds each, and evaluate all models
on the independently designed ModelNet40-C. Invariance training transfers
narrowly but reliably. It improves exactly the corruption families its
perturbations target wherever the baseline has headroom (mean occlusion $+10.5$, lidar $+13.4$ percentage points of accuracy). At matched training budget the best
standard backbone and the best RI architecture end on par
on unseen corruptions (within 2.5 percentage points in every
seed), and invariance-trained RIConv++ is the most
seed-stable top model. Once rotation augmentation is widened,
standard backbones concede only a flat, angle-independent accuracy
offset under full-sphere rotation. We release the benchmark along with a package for
analyzing point cloud robustness and reproducing our results.
\end{abstract}

\section{Introduction}
\label{sec:intro}

\begin{figure}[t]
\centering
\includegraphics[width=\linewidth]{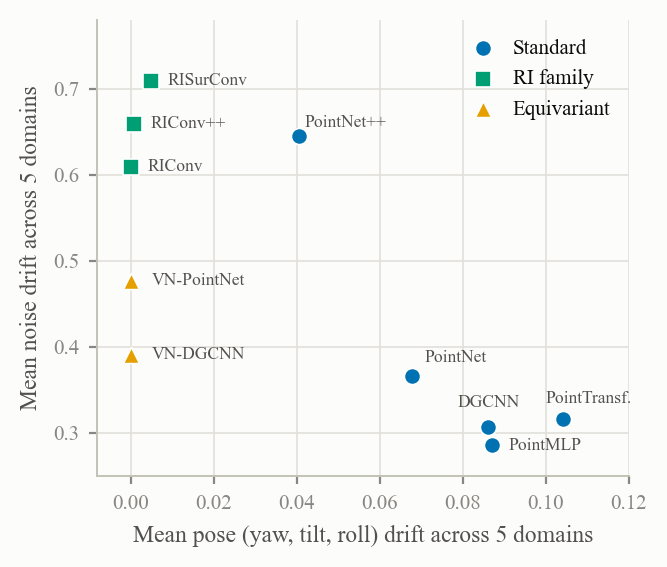}
\caption{The pose/noise trade-off measured in this paper. Each point
is one backbone: mean embedding drift $D$ across all 5 domains, with
pose pooled over the three single-axis rotations (yaw, tilt, roll;
ModelNet40 cells from replicate 1). The RI family (green squares) and
the equivariant VN models (orange triangles) drive pose drift to zero
but pay for it in noise. Standard backbones (blue circles) show the
opposite pattern. No backbone reaches the bottom-left corner.}
\label{fig:teaser}
\end{figure}

Point-based deep learning serves safety- and reliability-critical
applications such as autonomous driving, robotic manipulation, and
AR/VR scene understanding, where LiDAR and depth sensors produce
unordered, sparse, and frequently incomplete 3D point sets. Since PointNet~\cite{pointnet} first showed that a
permutation-invariant network operating directly on raw points could
rival voxel- and mesh-based approaches, the field has advanced rapidly.
PointNet++~\cite{pointnetpp} introduced hierarchical local feature
learning, DGCNN~\cite{dgcnn} dynamic feature-space graphs, Point
Transformer~\cite{pointtransformer} local self-attention, and
PointMLP~\cite{pointmlp} showed a residual MLP backbone can match more
elaborate designs. Almost all of this progress is trained and
benchmarked on clean, canonically aligned, synthetic CAD collections.

Real point clouds are not clean. Objects arrive under arbitrary
orientation, sensors inject positional noise, and self-occlusion or
clutter hides large parts of every surface. A substantial body of work
answers with architectures designed for one of these nuisances. The
most prominent is the rotation-invariant (RI) family built on local
reference frames and axes, with RIConv~\cite{riconv},
RIConv++~\cite{riconvpp}, and RISurConv~\cite{risurconv}, alongside
SO(3)-equivariant representations such as Vector
Neurons~\cite{vectorneurons}. These works typically report robustness
at a handful of rotation angles or a single noise setting, usually on
one dataset, and rarely with a statistical test attached.

This paper addresses two questions. First, what does robustness by
architecture actually cost? We measure this with a benchmark of 10
architectures, 5 domains, and 3 perturbation types, powered and
analyzed with confirmatory hypothesis tests and exploratory ANOVA.
The answer is a pose-versus-noise trade-off. The RI family
achieves the best pose robustness, and the rotation drift of RIConv
and RIConv++ is statistically indistinguishable from zero. The same family ranks among the lowest
on noise robustness in 4 of 5 domains, with effect sizes above 1.
Second, is that cost intrinsic to invariance, or only to obtaining it
architecturally? We retrain all 10 architectures unchanged,
with a simple cross-perturbation invariance loss on the embedding,
and evaluate all 20 resulting models on ModelNet40-C, a corruption
benchmark designed independently of our training perturbations. The
result reframes the trade-off. Invariance training improves exactly
the corruption families its perturbations target, in all 10 architectures, wherever the baseline has headroom, and it does not
transfer across families. An augmentation-only control traces these
gains to the perturbed views; the loss limits the regressions that
augmentation alone causes on other families. Under this regime the best standard
backbone and the best RI architecture end on par on
unseen corruptions across training seeds. The RI family's remaining
advantage reduces to exact pose invariance, which we quantify with
rotation sweeps and nearly close with widened rotation augmentation.

Our contributions are:
\begin{itemize}
  \item A statistically powered robustness benchmark of 10 point
  cloud architectures across 5 domains and 3 perturbation types, with
  confirmatory hypothesis tests, effect sizes, and exploratory ANOVA,
  released in full with a package for analyzing robustness and
  reproducing our results.
  \item A controlled isolation of the rotation-invariance versus noise
  robustness trade-off within architectures at matched statistical
  power, including the finding that exact architectural invariance is
  real (RIConv, mean rotation drift $-3\times10^{-4}$) and that it
  carries a measurable noise cost.
  \item A 20-model study of a cross-perturbation invariance loss,
  showing a narrow-transfer law: gains concentrate where a training
  perturbation matches the corruption and the baseline has headroom, and do not cross corruption families.
  \item Evidence that under matched invariance training the best
  standard backbone and the best RI architecture are on par
  on unseen corruptions across three training seeds. The
  concession to architectural invariance is quantified as a function
  of rotation angle, and a head-independent probe shows the gains
  belong to the representation rather than the trained head.
\end{itemize}

We present the invariance-loss study as the first scientific question
this benchmark can adjudicate.

\section{Related Work}
\label{sec:related}

\paragraph{Point cloud architectures.} PointNet~\cite{pointnet} and
PointNet++~\cite{pointnetpp} established point-wise MLPs with global
and hierarchical aggregation. DGCNN~\cite{dgcnn} recomputes graphs in
feature space, PointTransformer~\cite{pointtransformer} applies local
attention, and PointMLP~\cite{pointmlp} uses no sophisticated local
geometric extractors. The rotation-invariant family replaces raw coordinates with
handcrafted invariant features built on local reference axes:
RIConv~\cite{riconv}, RIConv++~\cite{riconvpp}, and the
attention-augmented RISurConv~\cite{risurconv}. Vector
Neurons~\cite{vectorneurons} instead lift scalar features to
equivariant 3D vectors (survey: \cite{rotationsurvey}).

\paragraph{Corruption benchmarks.} ModelNet40-C~\cite{modelnet40c}
applies 15 corruption types at 5 severities. PointCloud-C~\cite{pointcloudc}
simplifies the taxonomy to seven atomic corruptions and, via ShapeNet-C,
adds part segmentation. These benchmarks evaluate corruption
types in isolation, report point estimates without significance tests,
and do not cover the RI family. Ours attaches confirmatory tests and
effect sizes to every claim, covers the RI family, and isolates a
trade-off within architectures rather than ranking them.

\paragraph{Robustness interventions and narrow transfer.} In 2D,
robustness gained from training against one corruption famously fails
to generalize to others~\cite{geirhos2018,ford2019}. In 3D,
RobustPointSet~\cite{robustpointset} reports that robustness to one
transformation rarely survives train/test mismatch, which is the
closest precursor of our transfer profile. ModelNet40-C's authors evaluate
augmentation and test-time adaptation strategies on a small set of
backbones~\cite{modelnet40c}. Closer still, manifold
distillation~\cite{manifolddistillation} trains rotation-perturbation
robustness into a student network on the same four domains we share,
but it targets rotation alone and reports point estimates without a
transfer law or significance testing. None of the three measures an
embedding-level invariance loss across an architecture-perturbation
grid, and none conditions transfer on baseline headroom. Our finding
is the 3D, embedding-loss analog of the 2D narrow-transfer law, with
its conditions made explicit.

\section{Benchmark Design}
\label{sec:design}

\paragraph{Domains.} We use ModelNet40~\cite{modelnet40} (synthetic
CAD), ShapeNet~\cite{shapenet} (synthetic CAD), ABO~\cite{abo}
(product scans), ScanObjectNN~\cite{scanobjectnn} (real indoor
scans), and S3DIS~\cite{s3dis} (real building scans), giving two
synthetic and three real-world domains. All inputs are unit-sphere normalized
1024-point samples. Each domain uses a fixed 10-class subset (7 for
S3DIS, whose object-level classes are sparser), chosen for balanced
instance counts rather than to match any prior paper's split. The
subsets also contain no rotationally degenerate categories (bottle,
vase, cone), for which a yaw rotation is a near-identity and pose
effects would be understated. Our conclusions therefore concern
non-degenerate object categories.
Table~\ref{tab:domains} reports exact per-domain class, gallery,
query, and confirmatory pool sizes. The confirmatory pool powers
every (architecture, domain, perturbation) cell in
Sec.~\ref{sec:tradeoff}. Gallery is the training split, reused
unchanged as the reference set for the head-independent probe below.
Query is the held-out test split. The confirmatory pool is gallery
plus query minus one instance per class, withheld for the pilot
variance estimates and excluded from confirmatory testing to avoid
double-dipping.

\begin{table}[t]
\centering
\caption{Per-domain class counts and split sizes (instance counts, not
bytes). Confirmatory pool $=$ Gallery $+$ Query $-$ Classes (one
pilot instance per class withheld).}
\label{tab:domains}
\small
\setlength{\tabcolsep}{3.5pt}
\begin{tabular}{lcccc}
\toprule
Domain & Classes & Gallery & Query & Confirm. pool \\
\midrule
ModelNet40 & 10 & 450 & 300 & 740 \\
ShapeNet & 10 & 450 & 250 & 690 \\
ScanObjectNN & 10 & 200 & 120 & 310 \\
ABO & 10 & 200 & 120 & 310 \\
S3DIS & 7 & 140 & 84 & 217 \\
\bottomrule
\end{tabular}
\end{table}

\paragraph{Perturbations.} Pose is a rotation of the query relative to
the gallery, evaluated per axis (yaw, tilt, roll) at angle levels
$\{20, 40, 60, 80, 120\}$ degrees. The released generator additionally
produces a compound equal-magnitude three-axis rotation. We exclude it
from all reported statistics because its effective (geodesic) rotation
angle is non-monotonic in the nominal angle and peaks near 90 degrees,
so its severity levels are not ordered. Noise is additive Gaussian
positional noise with
$\sigma \in \{0.005, 0.01, 0.02, 0.03, 0.05\}$. Occlusion uses Hidden
Point Removal~\cite{katz2007hpr} from a random camera direction, with
severity controlled by a camera distance multiplier
$\in \{0.3, 0.6, 1.2, 2.5, 6.0\}$ of the cloud's bounding-box
diagonal. HPR operates
directly on points, so the same generator applies uniformly to every
domain, including real scans with no clean meshes. Severity ordering
is verified empirically in App.~\ref{sec:supp-pilot}.

\paragraph{Protocol and metrics.} Each architecture is trained per
domain, then evaluated on held-out queries. We report two levels.
Embedding drift $D$ is the per-instance drop in cosine similarity
between a fixed reference severity and the harshest severity (pose: 80
vs.\ 120 degrees; noise: $\sigma=0.005$ vs.\ $0.05$; occlusion: camera
multiplier 0.3 vs.\ 6.0). We aggregate it as mean $D$ with
paired one-sided t tests against zero, Cohen's $d_z$ effect sizes, and
confidence intervals. Classification accuracy is measured on corrupted
inputs. Both levels matter and can disagree (Sec.~\ref{sec:rotation}
shows one decisive case). App.~\ref{sec:supp-commensurability}
corroborates the cross-architecture drift orderings with accuracy
analogs (Spearman $\rho$ of 0.90 to 0.98 across the three
perturbation types).

\paragraph{Statistical machinery.} Each (architecture, domain,
perturbation) cell carries three pre-registered hypothesis families,
none corrected against each other. The primary test asks whether the
perturbation reduces embedding similarity at all: a one-sided paired
t test of mean drift against zero at $\alpha=0.05$. It is powered for
a fixed a priori minimal interesting effect $\delta=0.05$ in raw
cosine-similarity units at 80\% target power. Required $n$ is solved
per domain from pilot-estimated variance via a noncentral-$t$
calculation, which is why the confirmatory pool sizes in
Table~\ref{tab:domains} differ by domain (217 to 740).
App.~\ref{sec:supp-pilot} validates the pilots against realized
variance (every accepted non-rejection powered $\geq0.97$, one cell
inconclusive). A one-sided Wilcoxon signed-rank test is reported
alongside every primary test as a distribution-free cross-check
(Shapiro-Wilk routinely rejects normality of the drift distribution,
$p<10^{-4}$ in most cells). The secondary test asks whether severity
level itself matters, across the 5 severities defined above for each
perturbation type. It is a randomized-block one-way ANOVA with
Fligner-Killeen and Shapiro-Wilk diagnostics on its residuals,
reported alongside a Friedman test as an unconditional non-parametric
fallback rather than a contingency. When the omnibus Friedman test
rejects at $\alpha=0.05$, we run Bonferroni-corrected pairwise
Wilcoxon post-hoc tests across all 10 severity pairs
($\alpha_{\mathrm{adj}}=\alpha/10=0.005$). A third test asks whether
embedding drift differs across the architectures within a domain,
again a Friedman test at $\alpha=0.05$. We do not correct across the
250 primary cells (10 backbones, the 8 architectures above plus the
equivariant VN-PointNet and VN-DGCNN, $\times$ 5 domains $\times$ 5
perturbation variants: three rotation axes, noise, occlusion). Each is
powered and reported as its own pre-registered test
(Sec.~\ref{sec:tradeoff}). To state the scope precisely: the power
analysis backs per-cell existence and severity claims, while
cross-architecture comparisons rest on the Friedman test, effect
sizes, and App.~\ref{sec:supp-ci}'s confidence intervals rather than
on powered pairwise tests.

\paragraph{Head-independent probe.} All backbones in our suite end in
an L2-normalized embedding read by a linear classifier, a convention we
impose uniformly so head design is never a confound. To separate
representation quality from the trained head entirely, the benchmark
also ships a training-free probe: per-class subspaces are spanned by
the top singular vectors of the class's clean gallery-embedding matrix
(no mean centering, dimension 20) and a query is classified by
projection length, in the classical subspace-method
sense~\cite{oja1983,fukui2021subspace,msm}. We corroborate with cosine kNN ($k=5$,
similarity-weighted) and nearest-class-mean probes on the same frozen
embeddings, and with a nearest-class-mean control in raw Euclidean
space on pre-normalization features (full protocol in
App.~\ref{sec:supp-probe}). Agreement across heads licenses claims
about the embedding itself rather than any classifier.

\paragraph{Artifact.} We release the benchmark suite, the perturbation
and corruption generators, the statistical pipeline, the probe tools,
and all trained checkpoints, including the five widened-rotation-augmentation
retrains introduced in Sec.~\ref{sec:rotation}. Code is released
under the MIT license, checkpoints and results under CC~BY~4.0, and
no source data is redistributed (App.~\ref{sec:supp-artifact}). Project page:\\
{\Urlmuskip=0mu\url{https://abdullah-nasir-chowdhury.github.io/projects/wacv2027/}}.

\section{Results I: The Trade-off}
\label{sec:tradeoff}

\paragraph{Exact invariance is real.} On ModelNet40 yaw rotation
(replicate 1 of the seed study, App.~\ref{sec:supp-seeds}), RIConv's
and RIConv++'s mean embedding drifts are $-3\times10^{-4}$ and
$-5\times10^{-4}$ and the confirmatory tests fail to reject zero drift
($p=0.97$, $p=0.76$). RISurConv drifts at 0.002, an order of magnitude
below PointMLP (0.024), the smallest of the standard backbones
(PointNet 0.259,
DGCNN 0.169, PointTransformer 0.092, PointNet++ 0.066). VN-PointNet and
VN-DGCNN match this exactness in all 30 of their pose cells across
domains ($|D| < 10^{-4}$, App.~\ref{sec:supp-ci}). The RI
family's headline claim survives adversarially powered testing.

\paragraph{Noise cost.} Under noise the ordering inverts.
Table~\ref{tab:noise} shows mean noise drift per domain. On ModelNet40,
RIConv++ (0.537) and RISurConv (0.529) are the two most fragile of all
ten, and RIConv (0.447) drifts more than every standard backbone
except PointNet++ (0.454). All three RI cells have $d_z \geq 1.27$,
against PointNet at 0.76 and DGCNN at 1.09. The same inversion holds
on ShapeNet, ABO, and ScanObjectNN. On
S3DIS the ranking compresses and PointNet++ is the most fragile
(0.880). Noise fragility of the RI family generalizes across 4 of 5
domains. The equivariant route pays it too: despite exact pose
invariance, VN-PointNet's noise drift spans 0.332--0.619 across
domains and VN-DGCNN's 0.154--0.626. Occlusion drift shows no comparably consistent
family-level pattern across domains, and we make no family claim for
it.

\begin{table}[t]
\centering
\caption{Noise robustness as mean embedding drift $D$ (lower is
better). All cells $p<10^{-3}$. VN-PointNet and VN-DGCNN are
equivariant, not rotation-invariant (M40 column: replicate 1,
App.~\ref{sec:supp-seeds}; other domains: original runs). Per-cell
95\% CIs, here and for pose and occlusion, are in
App.~\ref{sec:supp-ci}.}
\label{tab:noise}
\small
\setlength{\tabcolsep}{3pt}
\begin{tabular}{lccccc}
\toprule
$D\downarrow$ & M40 & ShapeNet & ABO & ScanObj & S3DIS \\
\midrule
PointNet & 0.143 & 0.170 & 0.268 & 0.485 & 0.763 \\
PointNet++ & 0.454 & 0.333 & 0.712 & 0.844 & 0.880 \\
DGCNN & 0.105 & 0.250 & 0.230 & 0.367 & 0.583 \\
RIConv & 0.447 & 0.582 & 0.647 & 0.712 & 0.655 \\
RIConv++ & 0.537 & 0.561 & 0.673 & 0.774 & 0.752 \\
RISurConv & 0.529 & 0.671 & 0.764 & 0.744 & 0.836 \\
PointMLP & 0.264 & 0.084 & 0.331 & 0.388 & 0.359 \\
PointTransf. & 0.149 & 0.122 & 0.224 & 0.428 & 0.659 \\
VN-PointNet & 0.332 & 0.362 & 0.495 & 0.576 & 0.619 \\
VN-DGCNN & 0.154 & 0.313 & 0.325 & 0.536 & 0.626 \\
\bottomrule
\end{tabular}
\end{table}

\begin{figure*}[t]
\centering
\includegraphics[width=0.92\textwidth]{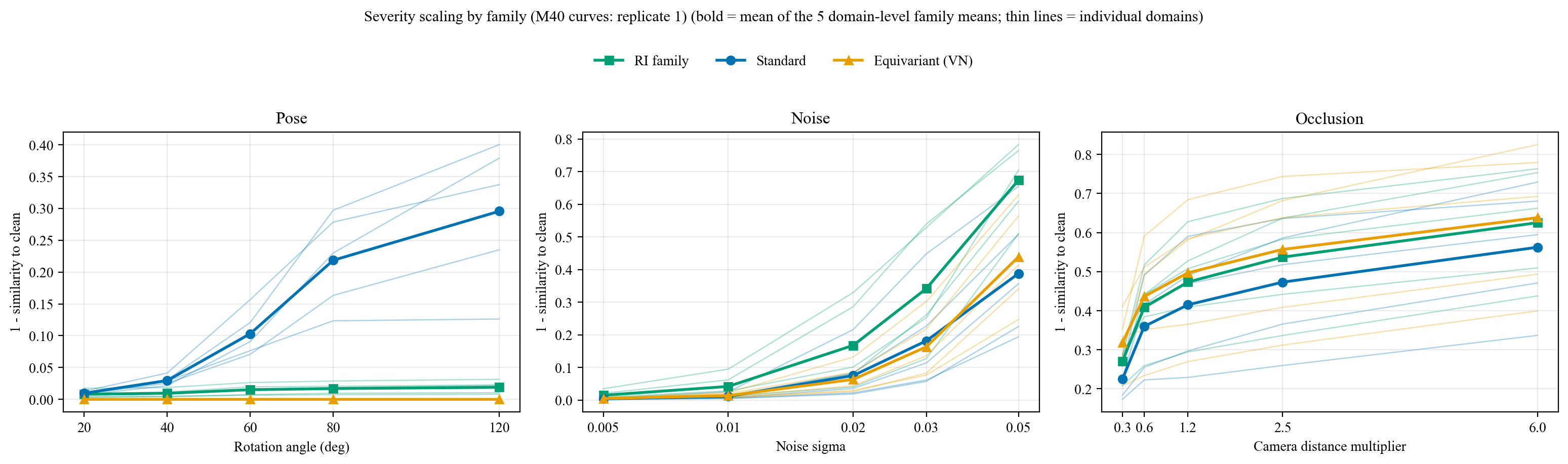}
\caption{Severity scaling for the standard (blue), RI (green), and
equivariant (orange; VN-PointNet and VN-DGCNN) families, with pose
pooled over yaw, tilt, and roll. ModelNet40 curves are from replicate
1. Lower is better. Bold lines are the mean of the 5 domain-level
family means, and thin lines are individual domains. The RI and
equivariant families' advantage on pose and
disadvantage on noise hold at every severity level, not just the
endpoints of Table~\ref{tab:noise}. Occlusion shows no comparable
family-level pattern: the families' domain lines interleave.}
\label{fig:severity}
\end{figure*}

\paragraph{Severity scaling.} The secondary severity test rejects
severity-independence in all 50 noise cells ($F>149$) and all 50
occlusion cells ($F>15$), at $p<10^{-4}$ with Friedman agreeing
throughout, and in 105 of the 150 per-axis pose cells. The 45 cells
that fail are exactly those of RIConv and the two VN models, whose
flat responses across angle levels restate their built-in invariance
(App.~\ref{sec:supp-vnseverity}). The trade-off is not an artifact of
one severity choice (Fig.~\ref{fig:severity}).

\paragraph{Signed drift.} $D$ is signed and the primary test is
one-sided. PointNet and DGCNN drift significantly \emph{negative} on
ModelNet40 roll ($-0.210$ and $-0.021$), so signed pooling flatters
their pose positions in
Fig.~\ref{fig:teaser}. Unsigned pooling would only widen the family
separation (App.~\ref{sec:supp-ci}).

The trade-off measures what built-in invariance costs, not whether
that cost is intrinsic. All three mechanisms tested here pay it:
noise-sensitive local reference axes (the RI family), equivariant
features (the VN models), and test-time PCA canonicalization, which
carries the same cost (App.~\ref{sec:supp-canon}). The rest of the
paper asks what happens when invariance is instead asked of the
training objective.

\section{The Intervention: Invariance by Training}
\label{sec:intervention}

We retrain each of the 10 architectures (the 8 above plus both VN
backbones), leaving the architecture
unchanged, end to end, with one added loss term. For a clean batch
$x$ with three perturbed views $T_r(x)$ (uniform SO(3) rotation),
$T_n(x)$ (Gaussian noise), and $T_o(x)$ (HPR occlusion),

\begin{equation}\label{eq:invloss}
\mathcal{L} = \mathrm{CE}(f(x), y) + \lambda \sum_{k \in \{r,n,o\}}
\left(1 - \cos\big(z(x), z(T_k(x))\big)\right)
\end{equation}

with $z$ the L2-normalized embedding and $\lambda = 0.5$. Training
perturbations deliberately reuse the benchmark's own generators, so
the internal evaluation measures matched-perturbation performance and
only ModelNet40-C measures generalization. Budget, optimizer, schedule,
and checkpoint selection are identical across all 20 models
(App.~\ref{sec:supp-training} for the full spec). The occlusion term
is computed on a sub-batch of 4 to reduce cost. We note this because
occlusion is where the loss gives the largest gains.

On the matched internal perturbations the loss achieves its
objective, and invariance-trained RIConv++ leads the internal table.
That is performance on trained-against perturbations. The question is
what generalizes.

\section{Results II: Generalization to Unseen Corruptions}
\label{sec:generalization}

We evaluate all 20 models (10 architectures under both training
conditions, replicate 1 of the three-seed study of
App.~\ref{sec:supp-seeds}) on ModelNet40-C~\cite{modelnet40c}, which
applies 15 corruption types at 5 severities and was designed
independently of our training perturbations. Because our models are trained on a 10-class
subset, mCE is recomputed internally with DGCNN as reference, and our
numbers are not comparable to the published ModelNet40-C leaderboard.

\begin{table}[t]
\centering
\caption{ModelNet40-C generalization, all 20 models (replicate 1; seed
spread: App.~\ref{sec:supp-seeds}, intervals:
App.~\ref{sec:supp-bootstrap}). mCE normalized to DGCNN baseline
($=1$); Gap is clean minus mean corrupted accuracy. The VN backbones
are equivariant, not rotation-invariant.}
\label{tab:m40c}
\footnotesize
\setlength{\tabcolsep}{3.5pt}
\begin{tabular}{llcccc}
\toprule
Arch & Loss & Clean$\uparrow$ & Corr.$\uparrow$ & mCE$\downarrow$ & Gap$\downarrow$ \\
\midrule
PointNet & no & 0.899 & 0.663 & 2.062 & 0.236 \\
PointNet++ & no & 0.890 & 0.756 & 1.545 & 0.134 \\
DGCNN & no & 0.934 & 0.806 & 1.000 & 0.128 \\
RIConv & no & 0.910 & 0.732 & 1.441 & 0.178 \\
RIConv++ & no & 0.901 & 0.775 & 1.194 & 0.126 \\
RISurConv & no & 0.892 & 0.691 & 1.892 & 0.201 \\
PointMLP & no & 0.923 & 0.762 & 1.241 & 0.161 \\
PointTransf. & no & 0.931 & 0.778 & 1.246 & 0.153 \\
VN-PointNet & no & 0.877 & 0.601 & 2.496 & 0.276 \\
VN-DGCNN & no & 0.874 & 0.646 & 2.171 & 0.228 \\
\midrule
PointNet & yes & 0.935 & 0.766 & 1.309 & 0.169 \\
PointNet++ & yes & 0.893 & 0.801 & 1.225 & \textbf{0.092} \\
DGCNN & yes & 0.941 & 0.788 & 1.200 & 0.153 \\
RIConv & yes & 0.917 & 0.807 & 1.070 & 0.110 \\
\textbf{RIConv++} & yes & 0.938 & \textbf{0.823} & \textbf{0.950} & 0.115 \\
RISurConv & yes & 0.753 & 0.609 & 2.639 & 0.144 \\
PointMLP & yes & 0.943 & \textbf{0.823} & 0.966 & 0.120 \\
PointTransf. & yes & 0.935 & 0.772 & 1.294 & 0.163 \\
VN-PointNet & yes & 0.883 & 0.601 & 2.521 & 0.282 \\
VN-DGCNN & yes & 0.891 & 0.606 & 2.517 & 0.285 \\
\bottomrule
\end{tabular}
\end{table}

\paragraph{The top of the table is a two-family tie.}
Invariance-trained RIConv++ and PointMLP are statistically tied for
the best mean corrupted accuracy of all 20 models (0.823 each, equivalent within $\pm 0.02$
by TOST, App.~\ref{sec:supp-bootstrap}). RIConv++ holds the
best mCE (0.950) and wins the rotation corruption type itself (0.900,
against 0.885 for the best invariance-trained standard backbone).
PointNet++ posts the smallest OOD gap (0.092). Across seeds
(App.~\ref{sec:supp-seeds}) the single best model rotates (PointMLP
here, PointNet++ in the other two replicates). RIConv++ is
top-two in every seed, on par with PointNet++ on corrupted accuracy
across seeds ($0.821 \pm 0.009$ vs.\ $0.823 \pm 0.020$), and it alone beats the
DGCNN mCE reference in every seed ($0.935 \pm 0.020$). After matched
training, what separates the families is not aggregate robustness but
exactness under rotation (Sec.~\ref{sec:tradeoff}) and training
stability (RISurConv, below).

\paragraph{The narrow-transfer law.} Aggregates hide the structure.
Table~\ref{tab:transfer} breaks the intervention's effect down by
corruption type. Gains concentrate almost entirely in two types, where all 10 architectures gain (RISurConv only marginally, below): lidar ($+13.4$ points on average) and occlusion
($+10.5$), where a point is one percentage point of accuracy.
Both are missing-surface corruptions matching the HPR training
perturbation, and both are where baselines fail hardest (36\% and 43\%
accuracy). Everything else moves within a few points of zero.
Training on the same views without the loss
(App.~\ref{sec:supp-augonly}) reproduces both gains but costs 5 to 6
points on rotation, shear, and distortion, where the loss stays within
1.3: the gains come from the views, the loss limits the cost.

\begin{table}[t]
\centering
\caption{Invariance-loss effect by corruption type: mean accuracy delta
(trained minus baseline) over the 10 architectures, with baseline mean
accuracy. Positive values indicate improvement. Types are grouped by
their relation to the training perturbations.}
\label{tab:transfer}
\small
\begin{tabular}{lcc}
\toprule
Corruption & $\Delta$ acc & Base acc \\
\midrule
\multicolumn{3}{l}{\emph{Trained family, headroom}} \\
lidar & \textbf{+0.134} & 0.364 \\
occlusion & \textbf{+0.105} & 0.434 \\
\midrule
\multicolumn{3}{l}{\emph{Trained family, no headroom}} \\
gaussian & +0.013 & 0.883 \\
cutout & +0.002 & 0.881 \\
rotation & $-$0.013 & 0.820 \\
\midrule
\multicolumn{3}{l}{\emph{Untrained family, headroom}} \\
background & $-$0.014 & 0.299 \\
\midrule
\multicolumn{3}{l}{\emph{Untrained family, no headroom}} \\
impulse & +0.058 & 0.696 \\
upsampling & +0.023 & 0.753 \\
uniform & +0.012 & 0.766 \\
density & $-$0.001 & 0.866 \\
density\_inc & $-$0.003 & 0.890 \\
distortion\_rbf & $-$0.007 & 0.790 \\
distortion\_rbf\_inv & $-$0.007 & 0.789 \\
shear & $-$0.011 & 0.794 \\
distortion & $-$0.011 & 0.788 \\
\bottomrule
\end{tabular}
\end{table}

The pattern is a conjunction. A corruption improves only when it
resembles a training perturbation and the baseline also has headroom.
Both falsification cells are present in the data. Cutout is
occlusion-family but mild (baseline 0.881) and gains nothing.
Background insertion has the most headroom of all types (0.299),
matches no training perturbation, and gains nothing. Rotation
and Gaussian noise are trained against, already easy, and gain
nothing. This is the 3D embedding-loss
analog of the 2D narrow-transfer results~\cite{geirhos2018,ford2019},
now measured across 10 architectures.

\paragraph{Architecture dependence is second order.} DGCNN's aggregate
mCE worsens under the loss (+0.200, significant,
App.~\ref{sec:supp-bootstrap}), yet DGCNN gains +0.064 on lidar and
+0.116 on occlusion like every other architecture. The aggregate loss
comes from small, consistent regressions across the distortion
family ($-0.05$ to $-0.07$ per type). One untested explanation is capacity reallocation toward the covered perturbation directions. RISurConv's collapse is a training-stability
problem (clean accuracy 0.892 to 0.753, recurring in all three
replicates, App.~\ref{sec:supp-seeds}) severe enough to suppress even
its target-corruption gains (+0.020 lidar, +0.007 occlusion). A warmup
arm (ramping $\lambda$ linearly from 0 to 0.5 over the first 10 epochs)
rescues RISurConv entirely, recovering clean accuracy above baseline
and mCE below 2. Run on all 8 original-study architectures, the
warmup helps five and hurts three, leaving this section's headline
unchanged (App.~\ref{sec:supp-warmup}). The plain run explains why:
validation accuracy sits at chance (0.10--0.19) through epoch 10
while the invariance term collapses toward zero almost immediately.
Training on the views without the loss collapses too
(App.~\ref{sec:supp-augonly}), so the trigger is the views from epoch
one, not the loss term; the warmup phases them in. This is schedule
sensitivity rather than an architectural limit.

\begin{figure}[t]
\centering
\includegraphics[width=\linewidth]{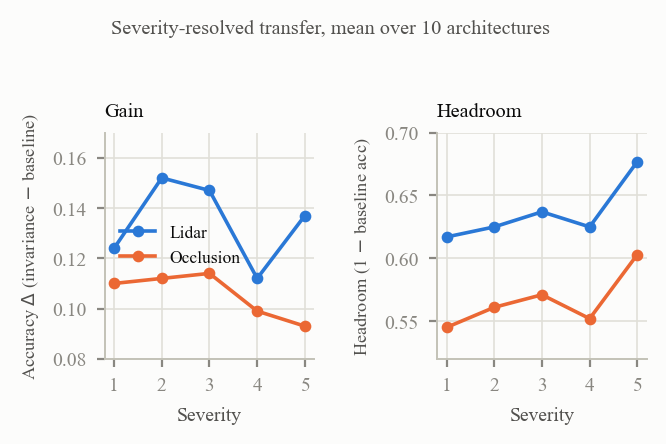}
\caption{Severity-resolved transfer, averaged over all 10
architectures: accuracy gain (left) and headroom (right) by
ModelNet40-C severity for the two target corruption types. Gains are
severity-stable. Headroom rises at the harshest severity without the
gains following.}
\label{fig:severity_resolved}
\end{figure}

\paragraph{Severity resolves the law.} Across ModelNet40-C's five
severities, headroom is a family-level fact, not a severity-level one. Both target types' gains
are severity-stable within a few points (occlusion $+0.093$ to
$+0.114$, lidar $+0.112$ to $+0.152$, neither monotone). Meanwhile
both types' headroom \emph{rises} at the harshest severity (to 0.603
and 0.677) without the gains following
(Fig.~\ref{fig:severity_resolved}). Headroom decides \emph{which} types gain, not how much within a type.

\paragraph{Head-independent confirmation.} The gains belong to the
representation, not the trained classifier. We extract frozen
embeddings for all 10 architectures and their
invariance-trained counterparts (replicate-1 checkpoints,
App.~\ref{sec:supp-seeds}) and
re-classify the same inputs with the three
training-free heads of Sec.~\ref{sec:design}.
Table~\ref{tab:probe-delta} (App.~\ref{sec:supp-probe}) gives the
result: occlusion gains stay
positive for nine of the ten architectures under every head (+12.3 to
+46.3 points; the exception is RISurConv, whose collapse is
head-independent, below), and noise gains stay positive everywhere
except DGCNN ($-3.3$ head, $-2.0$ subspace, $-1.7$ class-mean) and
VN-DGCNN's subspace head ($-4.7$). A Euclidean control
on the pre-normalization features attenuates the gains sharply
(e.g., PointNet's and PointNet++'s occlusion deltas), the expected signature of a
cosine loss that shapes embedding
direction, not magnitude, so only heads that read angular geometry see
the gain in full. Read App.~\ref{sec:supp-probe} for details.

\paragraph{RI fragility partly tracks training.}
RISurConv's clean-accuracy collapse (Sec.~\ref{sec:generalization})
shows up under every probe head, so it lives in the representation, not
the trained classifier, and what the loss still buys it on noise
shrinks to +0.3 to +6.7 points (App.~\ref{sec:supp-probe}).
RIConv++, invariant like RISurConv but under the same loss, shows what
the family can do when training doesn't collapse it: +45.0 to +52.7 on
noise, the largest gain of any architecture here, with pose-120
accuracy essentially untouched. Sec.~\ref{sec:tradeoff} establishes the
RI family among the most noise-fragile of all ten; RIConv++ shows that part of that fragility tracks training, not invariance by construction. App.~\ref{sec:supp-interaction} tests the
architecture$\times$training interaction formally, and RISurConv
carries it: excluded, the pose interaction vanishes and the two
surviving RI architectures gain more
from the loss on noise than the standard family in every seed. On the internal matched-perturbation benchmark,
invariance-trained RIConv++ even beats invariance-trained PointNet++
(App.~\ref{sec:supp-probe} works the arithmetic through).

\section{How Much Rotation Does the Loss Buy?}
\label{sec:rotation}

The RI family's remaining advantage is the rotation column, and we
measure it precisely. We
rotate each evaluation instance in 10-degree steps through 360 degrees
about yaw, tilt, roll, and random SO(3) axes, and record two curves per
model: embedding drift angle
$\theta(\alpha) = \arccos \big(\cos(z(x), z(R_\alpha x))\big)$, and
classification accuracy at each angle. Drift alone is not enough: a
flat drift curve can still coexist with a 20-point accuracy loss at 120
degrees.

\begin{figure}[t]
\centering
\includegraphics[width=0.9\linewidth]{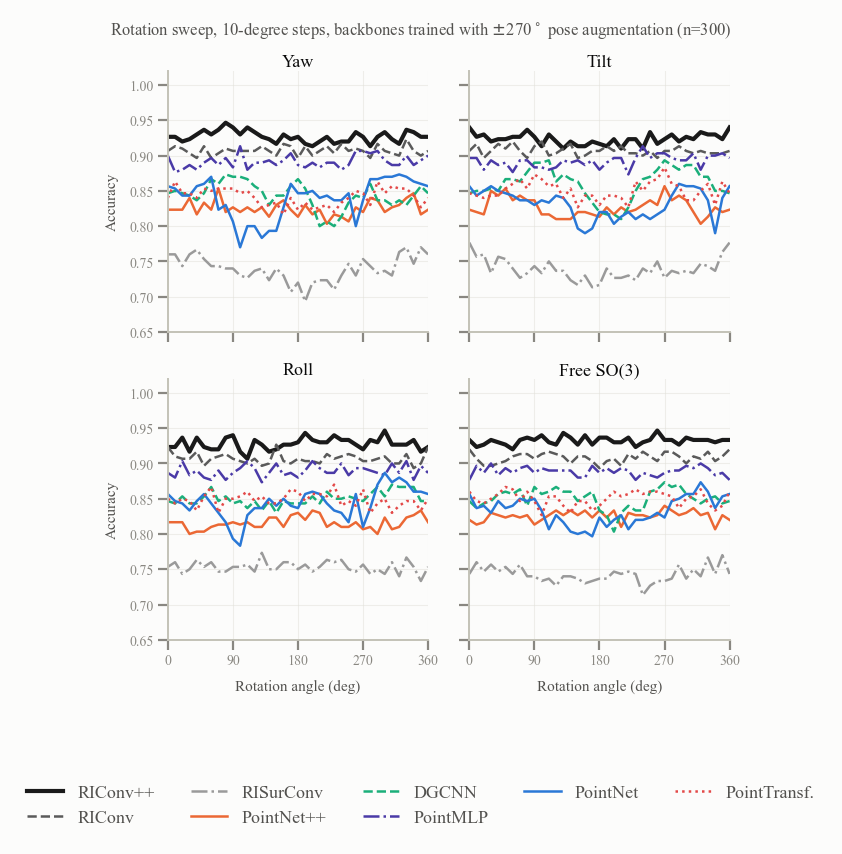}
\caption{Classification accuracy at 10-degree resolution through the
full 360-degree sweep, four axes, the five standard backbones
retrained with pose augmentation widened to $\pm270$ degrees about
yaw, tilt, and roll (all 300 held-out ModelNet40 test instances).
The RI family (grays) holds flat bands at every angle on every axis.
RIConv++ (black) and RIConv sit near the ceiling, while RISurConv is
flat but lower, reflecting the training collapse of
Sec.~\ref{sec:generalization}, not angle dependence.
The widened-augmentation backbones (colors) hold flat bands of their
own below the RI ceiling, showing offsets rather than troughs.
PointMLP tracks the RI band closest, DGCNN, PointTransformer, and
PointNet sit a few points lower, and PointNet++ pays the largest
offset.}
\label{fig:rotation_sweep}
\end{figure}

To ask how much of the remaining gap augmentation alone closes, we
retrain the five standard backbones with the training-time pose
augmentation widened to $\pm270$ degrees about yaw, tilt, and roll,
keeping the invariance loss and every other training choice fixed
(App.~\ref{sec:supp-training}: exact ranges, the z/SO(3) relation,
and that the loss's rotation views already sample SO(3)). We then
rotate
all 300 held-out ModelNet40 test instances (the full test split, 30
per class) in 10-degree steps through 360 degrees, about yaw, tilt,
roll,
and a free SO(3) axis (one random unit axis per instance). We record
classification accuracy for the three RI architectures and the five
widened-augmentation backbones at each of the resulting 144
(axis, angle) cells. Each
cell is a single forward pass per instance rather than the
5-replicate protocol of
Secs.~\ref{sec:design}--\ref{sec:generalization}. This is a
descriptive sweep sized for coverage rather than confirmatory power
(App.~\ref{sec:supp-training}). At $n=300$ each cell's accuracy
carries a standard error of 0.015 to 0.024, small enough to read the
per-angle structure directly.

Figure~\ref{fig:rotation_sweep} shows the result directly. RIConv++
holds a narrow band near ceiling, 0.907--0.947 accuracy, at every
sampled angle on every axis. The other two RI architectures, swept
under the identical protocol, are just as flat. RIConv's
invariance-loss model spans 0.893--0.927 across all 144 cells, and
RISurConv's spans 0.693--0.777, a flat band whose depressed level
traces to its documented training collapse
(Sec.~\ref{sec:generalization}), not angle dependence. The two
equivariant VN models, swept with their invariance-loss checkpoints,
are flatter still: accuracy is constant at 0.873 (VN-PointNet) and
0.843 (VN-DGCNN) in every one of the 144 cells. Flatness across
all five confirms built-in invariance is exact at 10-degree
resolution, not an artifact of the five coarse angles used elsewhere.
RIConv++ serves as
Table~\ref{tab:rotation_gap}'s RI reference because it is the
strongest of the three here. The widened-augmentation backbones sit
below that ceiling in flat bands of their own, not in any
angle-localized trough, and the offset is architecture-dependent
(Table~\ref{tab:rotation_gap}). PointMLP tracks the RI ceiling
closest, and PointNet++ carries the largest mean offset, which is the
across-the-board price of its own widened retrain.

\begin{table}[t]
\centering
\caption{Accuracy deficit against invariance-trained RIConv++ over
all 144 (axis, angle) cells, per widened-augmentation backbone and
for the two invariance-trained equivariant VN models under the same
sweep. The
deficit is present at the canonical pose itself and varies little with
angle: an offset, not a rotation effect
(App.~\ref{sec:supp-crossover}).}
\label{tab:rotation_gap}
\footnotesize
\begin{tabular}{lccc}
\toprule
Backbone & Mean gap$\downarrow$ & Widest gap$\downarrow$ & Cells $\geq0.05\downarrow$ \\
\midrule
PointMLP     & 0.037 & 0.067 & 0.16 \\
DGCNN        & 0.076 & 0.127 & 0.92 \\
PointTransf. & 0.080 & 0.113 & 0.98 \\
PointNet     & 0.091 & 0.160 & 0.98 \\
PointNet++   & 0.104 & 0.133 & 1.00 \\
VN-PointNet  & 0.054 & 0.073 & 0.65 \\
VN-DGCNN     & 0.084 & 0.103 & 1.00 \\
\bottomrule
\end{tabular}
\end{table}

The residual gap is not an extreme-rotation regime at all.
PointNet++'s deficit against RIConv++ is already 0.103 at zero
rotation, stays within 0.067 to 0.133 across all 144 cells, and
shows no far-from-canonical structure on any axis
(App.~\ref{sec:supp-crossover}). What remains is a flat,
angle-independent offset, the widened augmentation's own accuracy
cost spread over the whole sphere. That cost is real: widening the
augmentation to
$\pm270$ degrees pays for its pose coverage in clean accuracy (DGCNN
drops from 0.950 to 0.847) and in noise robustness (0.873 to 0.720 at
$\sigma=0.05$), the same trade this paper measures for architectural
invariance. Loss and augmentation are statistical while architectural
invariance is exact, but their practical difference is a flat offset,
listed per backbone in Table~\ref{tab:rotation_gap} and at most 6.7
points for the strongest backbone, PointMLP. The practitioner's
question is no longer whether the deployment reaches the far side of
the object. It is whether a few points of flat offset are worth wider
augmentation's clean and noise cost.

\section{Discussion}
\label{sec:discussion}

\paragraph{Do we still need rotation-invariant architectures?} At matched budget, robustness rank here no longer separates the families. Under
matched invariance training the best standard backbone and RIConv++
end on par on unseen corruptions across training seeds
(Sec.~\ref{sec:generalization}). What remains is a real choice on
other axes. A standard backbone avoids the RI family's training
instability (RISurConv's seed-recurrent collapse).
RIConv++, in return,
delivers exact pose invariance and the study's most seed-stable mCE.
What architectural invariance alone retains
is exactness with no augmentation tax. Applications that need that
guarantee (free-tumbling objects, arbitrary scan orientation) should
still pay for it.

\paragraph{Aim the loss.} The narrow-transfer law is practical
guidance. Training perturbations buy robustness to their own families
wherever the baseline is failing, and the loss limits the cost on the
others (App.~\ref{sec:supp-augonly}), so training families
should match the deployment's corruption profile. We see no consistent transfer to unmatched families.

\paragraph{Latent robustness.} On several
baselines the training-free probes outperform the model's own trained
classifier on corrupted inputs, a gap invariance training closes.
PointMLP at noise $\sigma=0.05$ is the extreme case: a 14.0-point gap between trained head and subspace
probe on the same frozen embedding (App.~\ref{sec:supp-probe}),
falling to 3.0 after invariance training. Corruption-robust class information can already
be present in a baseline embedding and simply go unused by its linear
head.

\paragraph{Limitations.} This is a controlled scientific comparison,
not a leaderboard result. Family claims cover the three RI and two equivariant architectures tested; our ModelNet40-C evaluation uses a 10-class subset with internally recomputed mCE, so only the 20-way comparison is claimed, not leaderboard-comparable numbers. Clean accuracies trail tuned references at this
budget: fragility here means fragility at matched budget. Statistical invariance carries no guarantee outside
the tested range, unlike architectural invariance. Sec.~\ref{sec:design}'s power
analysis covers evaluation-time variance only. The three-seed study
(App.~\ref{sec:supp-seeds}) shows the difference matters: family
effects reproduce across seeds, while single-model rankings do not.
$\lambda=0.5$ is a pre-committed convention
(App.~\ref{sec:supp-lambda} ablates it post hoc), and
App.~\ref{sec:supp-bootstrap} adds evaluation-set bootstrap intervals
to Sec.~\ref{sec:generalization}'s numbers (finite-test-set uncertainty only). Mapping how
transfer decays with dissimilarity to a held-out corruption family
is the natural next experiment.

\section{Conclusion}
\label{sec:conclusion}

We contribute a statistically powered robustness benchmark for point
cloud classification. We disentangle architectural
invariance from training-induced robustness, demonstrate that the
robustness trade-off reproduces across seeds and loss strengths,
and show that robustness transfer is selective, not universal. A simple invariance loss lifts the best standard and
RI models to parity on unseen corruptions across seeds. We release the benchmark,
generators, pipeline, and checkpoints.

{
    \small
    \bibliographystyle{ieeenat_fullname}
    \bibliography{main}

\begin{thebibliography}{25}
\providecommand{\natexlab}[1]{#1}
\providecommand{\url}[1]{\texttt{#1}}
\expandafter\ifx\csname urlstyle\endcsname\relax
  \providecommand{\doi}[1]{doi: #1}\else
  \providecommand{\doi}{doi: \begingroup \urlstyle{rm}\Url}\fi

\bibitem[Armeni et~al.(2016)Armeni, Sener, Zamir, Jiang, Brilakis, Fischer, and
  Savarese]{s3dis}
Iro Armeni, Ozan Sener, Amir~R. Zamir, Helen Jiang, Ioannis Brilakis, Martin
  Fischer, and Silvio Savarese.
\newblock {3D} semantic parsing of large-scale indoor spaces.
\newblock In \emph{CVPR}, 2016.

\bibitem[Chang et~al.(2015)Chang, Funkhouser, Guibas, Hanrahan, Huang, Li,
  Savarese, Savva, Song, Su, Xiao, Yi, and Yu]{shapenet}
Angel~X. Chang, Thomas Funkhouser, Leonidas Guibas, Pat Hanrahan, Qixing Huang,
  Zimo Li, Silvio Savarese, Manolis Savva, Shuran Song, Hao Su, Jianxiong Xiao,
  Li Yi, and Fisher Yu.
\newblock {ShapeNet}: An information-rich {3D} model repository.
\newblock \emph{arXiv preprint arXiv:1512.03012}, 2015.

\bibitem[Collins et~al.(2022)Collins, Goel, Deng, Luthra, Xu, Gundogdu, Zhang,
  Vicente, Dideriksen, Arora, Guillaumin, and Malik]{abo}
Jasmine Collins, Shubham Goel, Kenan Deng, Achleshwar Luthra, Leon Xu, Erhan
  Gundogdu, Xi Zhang, Tomas F.~Yago Vicente, Thomas Dideriksen, Himanshu Arora,
  Matthieu Guillaumin, and Jitendra Malik.
\newblock {ABO}: Dataset and benchmarks for real-world {3D} object
  understanding.
\newblock In \emph{CVPR}, 2022.

\bibitem[Deng et~al.(2021)Deng, Litany, Duan, Poulenard, Tagliasacchi, and
  Guibas]{vectorneurons}
Congyue Deng, Or Litany, Yueqi Duan, Adrien Poulenard, Andrea Tagliasacchi, and
  Leonidas Guibas.
\newblock Vector neurons: A general framework for {SO(3)}-equivariant networks.
\newblock In \emph{ICCV}, 2021.

\bibitem[Fei and Deng(2024)]{rotationsurvey}
Jiajun Fei and Zhidong Deng.
\newblock Rotation invariance and equivariance in {3D} deep learning: A survey.
\newblock \emph{Artificial Intelligence Review}, 57\penalty0 (7):\penalty0 168,
  2024.

\bibitem[Ford et~al.(2019)Ford, Gilmer, Carlini, and Cubuk]{ford2019}
Nicolas Ford, Justin Gilmer, Nicholas Carlini, and Ekin~D. Cubuk.
\newblock Adversarial examples are a natural consequence of test error in
  noise.
\newblock In \emph{International Conference on Machine Learning}, 2019.

\bibitem[Fukui(2021)]{fukui2021subspace}
Kazuhiro Fukui.
\newblock Subspace methods.
\newblock In \emph{Computer Vision: A Reference Guide}, pages 1221--1224.
  Springer, 2nd edition, 2021.

\bibitem[Geirhos et~al.(2018)Geirhos, Temme, Rauber, Sch\"utt, Bethge, and
  Wichmann]{geirhos2018}
Robert Geirhos, Carlos R.~M. Temme, Jonas Rauber, Heiko~H. Sch\"utt, Matthias
  Bethge, and Felix~A. Wichmann.
\newblock Generalisation in humans and deep neural networks.
\newblock In \emph{NeurIPS}, 2018.

\bibitem[Katz et~al.(2007)Katz, Tal, and Basri]{katz2007hpr}
Sagi Katz, Ayellet Tal, and Ronen Basri.
\newblock Direct visibility of point sets.
\newblock \emph{ACM TOG}, 2007.

\bibitem[Ma et~al.(2022)Ma, Qin, You, Ran, and Fu]{pointmlp}
Xu Ma, Can Qin, Haoxuan You, Haoxi Ran, and Yun Fu.
\newblock Rethinking network design and local geometry in point cloud: A simple
  residual {MLP} framework.
\newblock In \emph{ICLR}, 2022.

\bibitem[Oja(1983)]{oja1983}
Erkki Oja.
\newblock \emph{Subspace Methods of Pattern Recognition}.
\newblock Research Studies Press, Letchworth, England, 1983.

\bibitem[Qi et~al.(2017{\natexlab{a}})Qi, Su, Mo, and Guibas]{pointnet}
Charles~R. Qi, Hao Su, Kaichun Mo, and Leonidas~J. Guibas.
\newblock {PointNet}: Deep learning on point sets for {3D} classification and
  segmentation.
\newblock In \emph{CVPR}, 2017{\natexlab{a}}.

\bibitem[Qi et~al.(2017{\natexlab{b}})Qi, Yi, Su, and Guibas]{pointnetpp}
Charles~R. Qi, Li Yi, Hao Su, and Leonidas~J. Guibas.
\newblock {PointNet++}: Deep hierarchical feature learning on point sets in a
  metric space.
\newblock In \emph{NeurIPS}, 2017{\natexlab{b}}.

\bibitem[Ren et~al.(2022)Ren, Kong, Pan, and Liu]{pointcloudc}
Jiawei Ren, Lingdong Kong, Liang Pan, and Ziwei Liu.
\newblock {PointCloud-C}: Benchmarking and analyzing point cloud perception
  robustness under corruptions.
\newblock Preprint,
  {\Urlmuskip=0mu\url{https://pointcloud-c.github.io/asset/2022_PointCloud-C_preprint.pdf}},
  2022.

\bibitem[Sun et~al.(2022)Sun, Zhang, Kailkhura, Yu, Xiao, and Mao]{modelnet40c}
Jiachen Sun, Qingzhao Zhang, Bhavya Kailkhura, Zhiding Yu, Chaowei Xiao, and
  Z.~Morley Mao.
\newblock Benchmarking robustness of {3D} point cloud recognition against
  common corruptions.
\newblock \emph{arXiv preprint arXiv:2201.12296}, 2022.

\bibitem[Taghanaki et~al.(2020)Taghanaki, Luo, Zhang, Wang, Jayaraman, and
  Jatavallabhula]{robustpointset}
Saeid~Asgari Taghanaki, Jieliang Luo, Ran Zhang, Ye Wang, Pradeep~Kumar
  Jayaraman, and Krishna~Murthy Jatavallabhula.
\newblock {RobustPointSet}: A dataset for benchmarking robustness of point
  cloud classifiers.
\newblock \emph{arXiv preprint arXiv:2011.11572}, 2020.

\bibitem[Uy et~al.(2019)Uy, Pham, Hua, Nguyen, and Yeung]{scanobjectnn}
Mikaela~Angelina Uy, Quang-Hieu Pham, Binh-Son Hua, Thanh Nguyen, and Sai-Kit
  Yeung.
\newblock Revisiting point cloud classification: A new benchmark dataset and
  classification model on real-world data.
\newblock In \emph{ICCV}, 2019.

\bibitem[Wang et~al.(2019)Wang, Sun, Liu, Sarma, Bronstein, and Solomon]{dgcnn}
Yue Wang, Yongbin Sun, Ziwei Liu, Sanjay~E. Sarma, Michael~M. Bronstein, and
  Justin~M. Solomon.
\newblock Dynamic graph {CNN} for learning on point clouds.
\newblock \emph{ACM TOG}, 2019.

\bibitem[Wu et~al.(2015)Wu, Song, Khosla, Yu, Zhang, Tang, and
  Xiao]{modelnet40}
Zhirong Wu, Shuran Song, Aditya Khosla, Fisher Yu, Linguang Zhang, Xiaoou Tang,
  and Jianxiong Xiao.
\newblock {3D ShapeNets}: A deep representation for volumetric shapes.
\newblock In \emph{CVPR}, 2015.

\bibitem[Xu et~al.(2024)Xu, Liu, Wei, Xiong, Yu, and
  Zhang]{manifolddistillation}
Xinyu Xu, Huazhen Liu, Feiming Wei, Huilin Xiong, Wenxian Yu, and Tao Zhang.
\newblock Rotation perturbation robustness in point cloud analysis: A
  perspective of manifold distillation.
\newblock \emph{arXiv preprint arXiv:2411.01748}, 2024.

\bibitem[Yamaguchi et~al.(1998)Yamaguchi, Fukui, and Maeda]{msm}
Osamu Yamaguchi, Kazuhiro Fukui, and Ken-ichi Maeda.
\newblock Face recognition using temporal image sequence.
\newblock In \emph{IEEE International Conference on Automatic Face and Gesture
  Recognition}, 1998.

\bibitem[Zhang et~al.(2019)Zhang, Hua, Rosen, and Yeung]{riconv}
Zhiyuan Zhang, Binh-Son Hua, David~W. Rosen, and Sai-Kit Yeung.
\newblock Rotation invariant convolutions for {3D} point clouds deep learning.
\newblock In \emph{International Conference on 3D Vision (3DV)}, 2019.

\bibitem[Zhang et~al.(2022)Zhang, Hua, and Yeung]{riconvpp}
Zhiyuan Zhang, Binh-Son Hua, and Sai-Kit Yeung.
\newblock {RIConv++}: Effective rotation invariant convolutions for {3D} point
  clouds deep learning.
\newblock \emph{IJCV}, 2022.

\bibitem[Zhang et~al.(2024)Zhang, Yang, and Xiang]{risurconv}
Zhiyuan Zhang, Licheng Yang, and Zhiyu Xiang.
\newblock {RISurConv}: Rotation invariant surface attention-augmented
  convolutions for {3D} point cloud classification and segmentation.
\newblock In \emph{ECCV}, 2024.

\bibitem[Zhao et~al.(2021)Zhao, Jiang, Jia, Torr, and Koltun]{pointtransformer}
Hengshuang Zhao, Li Jiang, Jiaya Jia, Philip~H.S. Torr, and Vladlen Koltun.
\newblock Point transformer.
\newblock In \emph{ICCV}, 2021.

\end{thebibliography}
}

\clearpage
\appendix
\maketitlesupplementary

\section{Training Budget}
\label{sec:supp-training}

All 20 ModelNet40 models (10 architectures $\times$ \{baseline,
invariance loss\}, each retrained under three seeds,
App.~\ref{sec:supp-seeds}) are trained with one unified recipe,
verified against the training logs of the released checkpoints rather
than script defaults (invocation-time flags override some defaults):

\begin{itemize}
  \item \textbf{Optimizer}: Adam, learning rate $10^{-3}$,
  weight decay $10^{-4}$.
  \item \textbf{Schedule}: cosine annealing over the full run
  ($T_{\max}=$ total epochs).
  \item \textbf{Epochs}: 80, no early stopping (patience disabled).
  Every model trains the full 80 epochs regardless of accuracy
  plateaus.
  \item \textbf{Batch size}: effective 32 for every model. Two models
  reach it by gradient accumulation because their four-view forward
  pass under the loss (Eq.~\ref{eq:invloss}) exceeds GPU memory at true batch 32:
  PointMLP's invariance retrain (micro-batch 16 $\times$ 2) and
  VN-DGCNN's (micro-batch 8 $\times$ 4). Accumulated gradients are
  numerically identical to the big-batch step. The one behavioral
  difference is that batch normalization sees micro-batches, which we
  disclose rather than model.
  \item \textbf{Augmentation}: 20 augmented views per instance per
  epoch (repeats$=20$).
  \item \textbf{Checkpoint selection}: symmetric across conditions.
  Every model (baseline and invariance-loss alike) keeps the epoch
  with the highest accuracy on an internal 15\% stratified split of
  the gallery (split seed 42). The query split is never read during
  training or checkpoint selection.
  \item \textbf{Invariance loss} (retrains only): $\lambda=0.5$
  (Eq.~\ref{eq:invloss} in Sec.~\ref{sec:intervention}). The occlusion term of the loss is computed on a
  sub-batch of 4 instances per effective batch to reduce cost, the one
  asymmetry among the three loss terms. The sub-batch size is a fixed
  recipe constant, disclosed here and not ablated.
\end{itemize}

The cross-domain drift study (ShapeNet, ABO, ScanObjectNN, S3DIS) and
the warmup appendix use the original
study's checkpoints, whose recipe differed in two ways: baselines
selected on training accuracy (no validation split), and PointMLP's
invariance retrain ran at true batch 16. The canonicalization
(App.~\ref{sec:supp-canon}) and probe (App.~\ref{sec:supp-probe})
appendices use replicate 1 of the unified-recipe
checkpoints instead. The replicate study
reproduces that study's qualitative findings under the unified recipe
(App.~\ref{sec:supp-seeds}).

\paragraph{Widened-augmentation retrains (Sec.~\ref{sec:rotation}).}
The five models retrained with widened pose augmentation reuse the
invariance-loss recipe above unchanged except
for the training-time rotation augmentation ranges. The default
augmentation draws yaw from $\pm80$ degrees (steps of 20), pitch from
$\pm40$ (steps of 20), and roll from $\pm30$ (steps of 15), each step
perturbed by $\pm8$ degrees of uniform jitter. The widened variant
extends all
three axes to $\pm270$ degrees (steps of 30) with the same jitter.
Only the five standard backbones are retrained. For the RI family and
the equivariant VN models a pose-augmentation ablation is moot by
construction: their embeddings are exactly invariant to rotation
(Sec.~\ref{sec:tradeoff}), so rotated training views are feature-identical to the
originals.

Two protocol disclosures. First, the RI literature conventionally
reports z/z, z/SO(3), and SO(3)/SO(3) train/test settings. We widen
the benchmark's own per-axis stepped augmentation instead of
switching to a uniform-SO(3) draw because it extends the identical
generator the rest of the study trains with along one interpretable
dial (same steps-plus-jitter family, wider support). This keeps the
widened retrains comparable to every other model in the paper. The
evaluation side is not narrowed by this choice: the sweep covers full
360-degree ranges on all three canonical axes plus a free random
SO(3) axis. A uniform-SO(3)/SO(3) training arm is a straightforward
extension with the released generators. Second, ``augmentation
alone'' in Sec.~\ref{sec:rotation} means the marginal effect of
widening the \emph{data} augmentation: every model in that section,
including the default-augmentation ones, is trained with the
invariance loss, whose rotation view $T_r$ already draws uniform
SO(3) rotations (Eq.~\ref{eq:invloss}). Sec.~\ref{sec:rotation} therefore does not
decompose loss versus augmentation from scratch. It measures what
widening augmentation adds on top of a loss that already samples
SO(3). An augmentation-only arm at matched augmented-view budget
without the loss completes that decomposition on ModelNet40-C in
App.~\ref{sec:supp-augonly}.
The rotation sweep itself (Fig.~\ref{fig:rotation_sweep},
Table~\ref{tab:rotation_gap}) is one deterministic forward pass per
instance, axis, and angle, on the full 300-instance held-out test
split, for these five checkpoints and the RI family's replicate-1
invariance-loss checkpoints (App.~\ref{sec:supp-seeds}, matching the
provenance of every other ModelNet40 number in the paper), all swept
under the same protocol on the same instances. The RI architectures' forward passes batch the 36 angles
of each (instance, axis) pair into one call for tractability, with
every model in eval mode, so batch composition cannot affect
per-sample outputs.

\section{Warmup Schedule Ablation}
\label{sec:supp-warmup}

Sec.~\ref{sec:generalization} diagnoses RISurConv's collapse as
schedule sensitivity and rescues it with a $\lambda$ warmup (a linear
ramp from 0 at epoch 1 to 0.5 at epoch 11, held there afterwards). The same warmup arm was
trained for all 8 architectures under the otherwise identical
invariance-loss recipe and evaluated on ModelNet40-C under the
identical protocol. Table~\ref{tab:supp-warmup} reports all three
arms. All numbers in this section come from the original single-seed
recipe runs, so the loss column matches that study's arms rather than
Table~\ref{tab:m40c}'s replicate values. The schedule comparison is
within-recipe and unaffected. A single-seed warmup replicate under
the unified recipe, retrained for all 10 architectures during the
seed campaign, ships with the artifact. The warmup is not uniformly beneficial: it improves mCE for 5 of
8 architectures (PointNet 2.294 to 1.705, DGCNN 1.415 to 1.078,
RIConv++ 1.266 to 1.044, RISurConv 3.391 to 1.734, PointMLP 1.314 to
1.169) and worsens it for 3 (PointNet++ 1.007 to 1.126, RIConv 1.448
to 1.631, PointTransformer 1.483 to 1.854). Within the RI family the
RISurConv rescue does not generalize as a family property: RIConv++
posts its best arm under warmup, RIConv its worst. The headline
comparison of Sec.~\ref{sec:generalization} is unchanged by
best-of-both-schedules selection: no warmup arm reaches
invariance-trained PointNet++'s mean corrupted accuracy of 0.865 (the
closest are PointNet++'s own warmup arm at 0.854 and RIConv++'s at
0.842), and its 1.007 mCE remains the lowest of any loss-trained arm
under either schedule. Table~\ref{tab:m40c} reports the
identical-recipe arms only. The warmup arms are a schedule-sensitivity
ablation, not a third matched condition, and we do not mix schedules
within one table.

\begin{table}[t]
\centering
\caption{Warmup schedule ablation on ModelNet40-C, all 8
architectures: invariance-loss arm (identical recipe, as in the main
text) versus the same recipe with the $\lambda$ warmup. mCE is
internally normalized to the DGCNN baseline arm ($=1$), lower is
better.}
\label{tab:supp-warmup}
\small
\setlength{\tabcolsep}{3pt}
\begin{tabular}{lcccccc}
\toprule
 & \multicolumn{2}{c}{Clean$\uparrow$} & \multicolumn{2}{c}{Corr.$\uparrow$} & \multicolumn{2}{c}{mCE$\downarrow$} \\
\cmidrule(lr){2-3}\cmidrule(lr){4-5}\cmidrule(lr){6-7}
 & loss & warmup & loss & warmup & loss & warmup \\
\midrule
PointNet & 0.902 & 0.925 & 0.720 & 0.770 & 2.294 & 1.705 \\
PointNet++ & 0.920 & 0.934 & 0.865 & 0.854 & 1.007 & 1.126 \\
DGCNN & 0.952 & 0.945 & 0.796 & 0.828 & 1.415 & 1.078 \\
RIConv & 0.917 & 0.897 & 0.813 & 0.771 & 1.448 & 1.631 \\
RIConv++ & 0.935 & 0.949 & 0.817 & 0.842 & 1.266 & 1.044 \\
RISurConv & 0.743 & 0.904 & 0.624 & 0.762 & 3.391 & 1.734 \\
PointMLP & 0.940 & 0.935 & 0.804 & 0.829 & 1.314 & 1.169 \\
PointTransf. & 0.942 & 0.912 & 0.789 & 0.747 & 1.483 & 1.854 \\
\bottomrule
\end{tabular}
\end{table}

\section{Augmentation-Only Control}
\label{sec:supp-augonly}

The invariance loss trains against perturbed views, so its gains
could come from the loss term or from exposure to the views
themselves. To separate the two, we retrain all 10 architectures under
the replicate-1 recipe with the loss removed and its own three
generators applied as ordinary data augmentation. Each training view
is independently rotated (uniform SO(3)), noised, or occluded, or left
unperturbed, with probabilities 0.32, 0.32, 0.04, and 0.32. This is
the per-step view composition of the invariance-loss trainers, and the
number of dataset repeats is raised so that the perturbed-view budget
matches theirs (about 62.5 views per instance per epoch). Training uses
the classification loss only, with the same validation-split
checkpoint selection, and evaluation follows the identical
ModelNet40-C protocol. For the RI and VN models the rotation view
leaves the embedding unchanged (Sec.~\ref{sec:tradeoff}), but the noise and occlusion
views do not, so all 10 architectures receive the full control.
Table~\ref{tab:supp-w4} reports the three arms per architecture and
Table~\ref{tab:supp-w4-types} breaks the effect down by corruption
type.

\paragraph{Where the gains come from.} The gains on the targeted
missing-surface corruptions come from the perturbed views, not from
the loss. Augmentation alone improves lidar by $+0.170$ and occlusion
by $+0.123$ on average, at least as much as the loss ($+0.134$ and
$+0.105$), and improves both for 9 of the 10 architectures (RISurConv
is the exception). The two arms differ on the corruptions they do not
target. Augmentation alone loses 5.4 to 5.8 points on rotation, shear,
and the three distortion types, where the loss arm stays within 1.3
points of the baseline, and it moves upsampling by $-0.022$ against
the loss's $+0.023$. As a result, augmentation alone leaves mean
corrupted accuracy over the 10 architectures unchanged ($0.721$, as
for the baseline) while the loss arm raises it to $0.740$, and mean
mCE worsens under augmentation alone ($1.629 \to 1.746$) but improves
under the loss ($1.629 \to 1.569$). The concentration of gains in
lidar and occlusion (Sec.~\ref{sec:generalization}) holds under both
arms.

\paragraph{Per architecture.} The ranking of the two arms depends on
the architecture. Augmentation alone reaches higher corrupted accuracy
than the loss for PointNet++ (0.846 vs.\ 0.801), RIConv++ (0.829
vs.\ 0.823), and VN-PointNet (0.656 vs.\ 0.601). The loss arm is ahead
for the other seven, by 7.6 points for PointNet, 6.1 for PointMLP, and
4.5 for VN-DGCNN. RISurConv also degrades without the loss: its
augmentation-only arm reaches clean accuracy 0.732 and corrupted
accuracy 0.529, below its invariance-loss arm (0.753 and 0.609), so
the training instability discussed in Sec.~\ref{sec:generalization}
occurs under perturbed-view training even without the invariance term.

\paragraph{Scope.} These are single training runs per arm, so seed
variance of the size documented in App.~\ref{sec:supp-seeds} applies. In
particular, the augmentation-only PointNet++ and RIConv++ values
exceed the best value in the matched 20-model comparison
(Table~\ref{tab:m40c}), but they belong to a different training
condition and a single seed, and we do not rank them against that
table.

\begin{table}[t]
\centering
\caption{Augmentation-only control on ModelNet40-C, all 10
architectures: baseline, augmentation-only arm (the invariance loss's
rotation, noise, and occlusion views as data augmentation at matched
view budget, no invariance loss), and invariance-loss arm, all under
the replicate-1 recipe. Baseline and loss columns are those of
Table~\ref{tab:m40c}. mCE is normalized to the replicate-1 DGCNN
baseline ($=1$). One training run per arm.}
\label{tab:supp-w4}
\scriptsize
\setlength{\tabcolsep}{2.5pt}
\begin{tabular}{lccccccccc}
\toprule
 & \multicolumn{3}{c}{Clean$\uparrow$} & \multicolumn{3}{c}{Corr.$\uparrow$} & \multicolumn{3}{c}{mCE$\downarrow$} \\
\cmidrule(lr){2-4}\cmidrule(lr){5-7}\cmidrule(lr){8-10}
 & base & aug & loss & base & aug & loss & base & aug & loss \\
\midrule
PointNet & 0.899 & 0.930 & 0.935 & 0.663 & 0.690 & 0.766 & 2.062 & 1.902 & 1.309 \\
PointNet++ & 0.890 & 0.880 & 0.893 & 0.756 & 0.846 & 0.801 & 1.545 & 0.985 & 1.225 \\
DGCNN & 0.934 & 0.911 & 0.941 & 0.806 & 0.776 & 0.788 & 1.000 & 1.250 & 1.200 \\
PointMLP & 0.923 & 0.923 & 0.943 & 0.762 & 0.762 & 0.823 & 1.241 & 1.401 & 0.966 \\
PointTransf. & 0.931 & 0.938 & 0.935 & 0.778 & 0.766 & 0.772 & 1.246 & 1.390 & 1.294 \\
RIConv & 0.910 & 0.913 & 0.917 & 0.732 & 0.792 & 0.807 & 1.441 & 1.238 & 1.070 \\
RIConv++ & 0.901 & 0.949 & 0.938 & 0.775 & 0.829 & 0.823 & 1.194 & 0.979 & 0.950 \\
RISurConv & 0.892 & 0.732 & 0.753 & 0.691 & 0.529 & 0.609 & 1.892 & 3.209 & 2.639 \\
VN-PointNet & 0.877 & 0.897 & 0.883 & 0.601 & 0.656 & 0.601 & 2.496 & 2.205 & 2.521 \\
VN-DGCNN & 0.874 & 0.907 & 0.891 & 0.646 & 0.561 & 0.606 & 2.171 & 2.899 & 2.517 \\
\midrule
Mean & 0.903 & 0.898 & 0.903 & 0.721 & 0.721 & 0.740 & 1.629 & 1.746 & 1.569 \\
\bottomrule
\end{tabular}
\end{table}

\begin{table}[t]
\centering
\caption{Augmentation-only versus invariance-loss effect by corruption
type: mean accuracy delta (arm minus baseline) over the 10
architectures, with the number of architectures that improve in
parentheses. Grouping and the loss column as in
Table~\ref{tab:transfer}.}
\label{tab:supp-w4-types}
\small
\begin{tabular}{lcc}
\toprule
Corruption & $\Delta$ aug-only & $\Delta$ loss \\
\midrule
\multicolumn{3}{l}{\emph{Trained family, headroom}} \\
lidar & +0.170 (9) & +0.134 (10) \\
occlusion & +0.123 (9) & +0.105 (10) \\
\midrule
\multicolumn{3}{l}{\emph{Trained family, no headroom}} \\
gaussian & +0.013 (8) & +0.013 (8) \\
cutout & $-$0.008 (5) & +0.002 (7) \\
rotation & $-$0.054 (2) & $-$0.013 (4) \\
\midrule
\multicolumn{3}{l}{\emph{Untrained family, headroom}} \\
background & $-$0.019 (6) & $-$0.014 (5) \\
\midrule
\multicolumn{3}{l}{\emph{Untrained family, no headroom}} \\
impulse & +0.043 (6) & +0.058 (6) \\
upsampling & $-$0.022 (6) & +0.023 (6) \\
uniform & $-$0.017 (5) & +0.012 (5) \\
density & $-$0.006 (6) & $-$0.001 (7) \\
density\_inc & 0.000 (5) & $-$0.003 (7) \\
distortion\_rbf & $-$0.057 (2) & $-$0.007 (4) \\
distortion\_rbf\_inv & $-$0.055 (3) & $-$0.007 (4) \\
shear & $-$0.056 (3) & $-$0.011 (3) \\
distortion & $-$0.058 (2) & $-$0.011 (4) \\
\bottomrule
\end{tabular}
\end{table}

\section{Evaluation-Set Uncertainty on ModelNet40-C}
\label{sec:supp-bootstrap}

Table~\ref{tab:m40c} reports point estimates. This section attaches
evaluation-set uncertainty to every number in it: a nonparametric
bootstrap over the 1{,}000 evaluation instances (2{,}000 draws,
percentile 95\% intervals), with the same instance resample applied to
every model within a draw, so that paired contrasts inherit the
pairing. mCE is recomputed inside each draw, reference errors
included, making the interval on each ratio honest about numerator and
denominator alike. The DGCNN baseline, as the normalizer, has mCE 1 in
every draw by construction and carries no interval. These intervals
quantify uncertainty from the finite evaluation set only. Training-seed
variance is a distinct quantity with its own study
(Sec.~\ref{sec:discussion}), and the two are complementary: an
interval here says how firmly this checkpoint's numbers are measured,
not how firmly the training procedure reproduces them.

Table~\ref{tab:supp-bootstrap} gives the absolute intervals and
Table~\ref{tab:supp-bootstrap-delta} the paired invariance-loss
deltas, both for replicate 1. The seed appendix
(App.~\ref{sec:supp-seeds}) reports the spread of every number across
all three replicates. Three observations. First, the intervals
resolve the top of the table into a statistical tie: invariance-trained
PointMLP ($0.823$ $[0.810, 0.836]$) and invariance-trained RIConv++
($0.823$ $[0.809, 0.836]$) are indistinguishable on mean corrupted
accuracy, and RIConv++ posts the best mCE of any model ($0.950$, the
only arm whose interval reaches below the DGCNN reference alongside
PointMLP's $0.966$). Which single model sits first is
seed-dependent (App.~\ref{sec:supp-seeds}). The shared top is
not. Second, pairing is what buys resolution. Because both arms
of an architecture share every resample, the delta intervals are
several times tighter than the marginal ones, and five of the ten
corrupted-accuracy deltas are individually significant gains
(PointNet $+0.103$, RIConv $+0.075$, PointMLP $+0.062$, RIConv++
$+0.048$, PointNet++ $+0.045$), while RISurConv's collapse shows as a
significant $-0.082$. Third, the equivariant family is the one family
the loss does not help. VN-PointNet's delta is exactly zero
($[-0.012, +0.013]$) and VN-DGCNN's is significantly negative
($-0.040$, mCE $+0.345$), so the corruption fragility of the VN
backbones is not a trained-head artifact the loss can repair.

\paragraph{Equivalence, not just non-significance.} Overlapping
intervals show only that a difference is not detected. To claim the
headline tie positively we run two one-sided tests (TOST) on the same
2{,}000 paired draws. Equivalence at $\alpha=0.05$ requires the 90\%
percentile interval of the corrupted-accuracy difference to fall
inside $\pm 0.02$, the scale of seed-to-seed variation in the top
models' corrupted accuracy ($0.009$ to $0.023$,
App.~\ref{sec:supp-seeds}). A difference passes as negligible only if
it is confidently smaller than retraining noise. The headline pair
passes.
Invariance-trained RIConv++ and PointMLP differ by $-0.001$ with a
90\% interval of $[-0.009, +0.009]$, well inside the margin, so the
replicate-1 tie is demonstrated equivalence rather than absence of
evidence. The same test shows why we claim no per-seed winner between
RIConv++ and PointNet++. Their difference is $+0.022$ in replicate 1,
$-0.005$ in replicate 2, and $-0.023$ in replicate 3, significant in
opposite directions in the first and last. The training seed moves
this contrast across its own equivalence margin, which is the concrete
sense in which the tie is a family-level fact and single-model
rankings are not stable.

\begin{table*}[t]
\centering
\caption{Evaluation-set bootstrap intervals for Table~\ref{tab:m40c},
all 20 models (replicate 1): percentile 95\% intervals over 2{,}000
paired resamples of the 1{,}000 evaluation instances. The DGCNN
baseline is the mCE normalizer, so its mCE is 1 in every draw and
carries no interval. The two VN
backbones are equivariant, not rotation-invariant.}
\label{tab:supp-bootstrap}
\footnotesize
\setlength{\tabcolsep}{4pt}
\begin{tabular}{llccc}
\toprule
Arch & Loss & Clean$\uparrow$ & Corr.$\uparrow$ & mCE$\downarrow$ \\
\midrule
PointNet & no & 0.899 [0.879, 0.917] & 0.663 [0.646, 0.680] & 2.062 [1.844, 2.316] \\
PointNet++ & no & 0.890 [0.870, 0.909] & 0.756 [0.741, 0.769] & 1.545 [1.400, 1.714] \\
DGCNN & no & 0.934 [0.918, 0.948] & 0.806 [0.792, 0.820] & 1.000 (ref.) \\
RIConv & no & 0.910 [0.891, 0.927] & 0.732 [0.718, 0.745] & 1.441 [1.300, 1.612] \\
RIConv++ & no & 0.901 [0.882, 0.919] & 0.775 [0.761, 0.787] & 1.194 [1.080, 1.328] \\
RISurConv & no & 0.892 [0.873, 0.910] & 0.691 [0.675, 0.705] & 1.892 [1.713, 2.107] \\
PointMLP & no & 0.923 [0.906, 0.940] & 0.762 [0.748, 0.774] & 1.241 [1.129, 1.374] \\
PointTransf. & no & 0.931 [0.915, 0.946] & 0.778 [0.765, 0.790] & 1.246 [1.142, 1.372] \\
VN-PointNet & no & 0.877 [0.856, 0.897] & 0.601 [0.584, 0.618] & 2.496 [2.240, 2.816] \\
VN-DGCNN & no & 0.874 [0.851, 0.894] & 0.646 [0.628, 0.662] & 2.171 [1.964, 2.421] \\
\midrule
PointNet & yes & 0.935 [0.919, 0.950] & 0.766 [0.750, 0.782] & 1.309 [1.161, 1.484] \\
PointNet++ & yes & 0.893 [0.873, 0.911] & 0.801 [0.788, 0.813] & 1.225 [1.111, 1.360] \\
DGCNN & yes & 0.941 [0.926, 0.954] & 0.788 [0.774, 0.802] & 1.200 [1.099, 1.321] \\
RIConv & yes & 0.917 [0.900, 0.934] & 0.807 [0.794, 0.820] & 1.070 [0.958, 1.210] \\
RIConv++ & yes & 0.938 [0.923, 0.952] & \textbf{0.823} [0.809, 0.836] & \textbf{0.950} [0.846, 1.064] \\
RISurConv & yes & 0.753 [0.727, 0.779] & 0.609 [0.589, 0.629] & 2.639 [2.329, 3.003] \\
PointMLP & yes & 0.943 [0.929, 0.957] & \textbf{0.823} [0.810, 0.836] & 0.966 [0.872, 1.074] \\
PointTransf. & yes & 0.935 [0.919, 0.950] & 0.772 [0.758, 0.786] & 1.294 [1.187, 1.427] \\
VN-PointNet & yes & 0.883 [0.861, 0.902] & 0.601 [0.585, 0.617] & 2.521 [2.255, 2.848] \\
VN-DGCNN & yes & 0.891 [0.871, 0.909] & 0.606 [0.590, 0.622] & 2.517 [2.252, 2.837] \\
\bottomrule
\end{tabular}
\end{table*}

\begin{table}[t]
\centering
\caption{Paired invariance-loss deltas (trained minus baseline,
replicate 1; point estimates) with bootstrap intervals: because both arms share each
instance resample, these intervals are far tighter than the marginal
ones above. $^{*}$ marks intervals excluding zero.}
\label{tab:supp-bootstrap-delta}
\footnotesize
\setlength{\tabcolsep}{2.5pt}
\begin{tabular}{lcc}
\toprule
Arch & $\Delta$ Corr.$\uparrow$ & $\Delta$ mCE$\downarrow$ \\
\midrule
PointNet & $+$0.103 [$+$0.087, $+$0.120]$^{*}$ & $-$0.754 [$-$0.927, $-$0.601]$^{*}$ \\
PointNet++ & $+$0.045 [$+$0.035, $+$0.055]$^{*}$ & $-$0.320 [$-$0.415, $-$0.239]$^{*}$ \\
DGCNN & $-$0.018 [$-$0.030, $-$0.007]$^{*}$ & $+$0.200 [$+$0.099, $+$0.321]$^{*}$ \\
RIConv & $+$0.075 [$+$0.065, $+$0.086]$^{*}$ & $-$0.371 [$-$0.467, $-$0.282]$^{*}$ \\
RIConv++ & $+$0.048 [$+$0.039, $+$0.057]$^{*}$ & $-$0.244 [$-$0.335, $-$0.167]$^{*}$ \\
RISurConv & $-$0.082 [$-$0.098, $-$0.065]$^{*}$ & $+$0.747 [$+$0.570, $+$0.946]$^{*}$ \\
PointMLP & $+$0.062 [$+$0.053, $+$0.071]$^{*}$ & $-$0.274 [$-$0.362, $-$0.197]$^{*}$ \\
PointTransf. & $-$0.005 [$-$0.014, $+$0.003] & $+$0.048 [$-$0.027, $+$0.123] \\
VN-PointNet & $+$0.000 [$-$0.012, $+$0.013] & $+$0.025 [$-$0.081, $+$0.124] \\
VN-DGCNN & $-$0.040 [$-$0.056, $-$0.023]$^{*}$ & $+$0.345 [$+$0.207, $+$0.497]$^{*}$ \\
\bottomrule
\end{tabular}
\end{table}

\section{Architecture--Training Interaction}
\label{sec:supp-interaction}

Secs.~\ref{sec:intervention} and~\ref{sec:generalization} move the
training dial with the architecture family held fixed. The title's
question in statistical form is the interaction: does the improvement
bought by the invariance loss \emph{differ} between the standard and
RI families, per perturbation type? We test this as a
difference-in-differences on the 16-condition accuracy evaluation (300
paired queries): per architecture,
$\Delta_a = \mathrm{acc}_{\mathrm{loss}} - \mathrm{acc}_{\mathrm{base}}$
with the five levels of each perturbation type pooled, and
$\mathrm{DiD} = \overline{\Delta}_{\mathrm{standard}} -
\overline{\Delta}_{\mathrm{RI}}$. Inference runs at two levels that
answer different questions. A paired instance bootstrap (2{,}000
draws, one shared resample per draw) yields evaluation-set intervals,
which ask whether the interaction is real \emph{for these eight
backbones}. An exact
permutation over family labels (all 56 five-versus-three assignments)
asks the stronger question of whether the family line is special, or
an arbitrary 5--3 split of these backbones could show the same
contrast.

Table~\ref{tab:supp-did} reports both for replicate 1, and every
claim below is checked against all three seed replicates. At the
instance level the interaction is real and signed as the trade-off
predicts: the loss buys the standard family more than it buys the RI
family on pose ($\mathrm{DiD} = +0.050$ $[+0.029, +0.070]$; $+0.077$
and $+0.089$ in replicates 2 and 3, significant in all three) and on
clean inputs (similarly stable). Two checks bound the claim's
strength, and we state both. First, no family-label permutation is
significant for these headline contrasts (smallest $p = 0.071$ across
all replicates and perturbations): eight backbones establish the
interaction for these backbones, not for the families as populations.
Second, the pose and clean interactions are carried by RISurConv's
collapse (Sec.~\ref{sec:generalization}). Excluding it, pose falls to
$+0.007$ $[-0.012, +0.026]$ and the noise DiD flips decisively
negative: $-0.089$ $[-0.109, -0.069]$ in replicate 1, with $-0.075$
and $-0.031$ in replicates 2 and 3, significant in all three, and the
one contrast in this study whose family-label permutation reaches
significance ($p = 0.048$). That is, the two surviving RI
architectures gain
\emph{more} from the loss on noise than the standard family does in
every seed, which is the headroom pattern of
Sec.~\ref{sec:generalization} at
family scale. The supported reading is deliberately bounded. Training
substantially mitigates the RI family's observed noise fragility,
robustly across seeds, and mistrained it can induce fragility, as
RISurConv's seed-recurrent collapse shows. A causal claim that
architecture makes such fragility impossible or necessary is not
licensed by eight backbones and is not made.

\begin{table}[t]
\centering
\caption{Architecture$\times$training interaction on the 16-condition
accuracy evaluation (replicate 1, $n=300$ paired queries, perturbation
levels pooled per type). Top: per-architecture accuracy delta $\Delta$
(invariance loss minus baseline). The
equivariant VN backbones are reported but never pooled into either
family. Bottom: difference-in-differences (standard-family mean
$\Delta$ minus RI-family mean $\Delta$) with paired-bootstrap 95\%
intervals ($^{*}$ excludes zero), the exact family-label permutation
$p$ (56 splits), and the same DiD with RISurConv excluded.}
\label{tab:supp-did}
\scriptsize
\setlength{\tabcolsep}{2.5pt}
\begin{tabular}{lcccc}
\toprule
 & Clean & Pose & Noise & Occl. \\
\midrule
PointNet & $+$0.027 & $+$0.037 & $+$0.044 & $+$0.126 \\
PointNet++ & $+$0.017 & $+$0.016 & $+$0.083 & $+$0.129 \\
DGCNN & $+$0.000 & $+$0.015 & $-$0.005 & $+$0.169 \\
PointMLP & $+$0.013 & $+$0.020 & $+$0.072 & $+$0.266 \\
PointTransf. & $+$0.027 & $+$0.011 & $+$0.026 & $+$0.149 \\
RIConv & $+$0.010 & $+$0.010 & $+$0.108 & $+$0.127 \\
RIConv++ & $+$0.020 & $+$0.016 & $+$0.157 & $+$0.247 \\
RISurConv & $-$0.133 & $-$0.117 & $-$0.083 & $-$0.019 \\
VN-PointNet & $+$0.027 & $+$0.027 & $+$0.084 & $+$0.137 \\
VN-DGCNN & $-$0.013 & $-$0.013 & $+$0.000 & $+$0.192 \\
\midrule
Mean, standard & $+$0.017 & $+$0.020 & $+$0.044 & $+$0.168 \\
Mean, RI & $-$0.034 & $-$0.030 & $+$0.061 & $+$0.118 \\
\midrule
DiD & $+$0.051$^{*}$ & $+$0.050$^{*}$ & $-$0.017 & $+$0.049$^{*}$ \\
\quad 95\% CI & [.026, .075] & [.029, .070] & [$-$.036, .002] & [.021, .079] \\
\quad perm.\ $p$ & 0.179 & 0.089 & 0.786 & 0.500 \\
DiD w/o RISur. & $+$0.002 & $+$0.007 & $-$0.089$^{*}$ & $-$0.020 \\
\quad 95\% CI & [$-$.025, .028] & [$-$.012, .026] & [$-$.109, $-$.069] & [$-$.049, .011] \\
\quad perm.\ $p$ & 0.952 & 0.524 & 0.048 & 0.952 \\
\bottomrule
\end{tabular}
\end{table}

\section{Loss-Weight Sensitivity}
\label{sec:supp-lambda}

The invariance loss's weight is fixed at $\lambda=0.5$ throughout this
paper as a pre-committed convention, chosen before any benchmark
evaluation and never tuned on it. This section ablates that choice
post hoc: $\lambda \in \{0.1, 0.25, 0.5, 1.0, 2.0\}$ for three
backbones spanning both families, with one training run per grid
point. The three are PointNet++ (the paper's overall
winner), PointNet (the worst-case regressor under the loss), and
RIConv (the cheapest RI representative; RISurConv's schedule
sensitivity is covered by App.~\ref{sec:supp-warmup}). The
equivariant models are not ablated because the loss does not improve
their corrupted accuracy at the default weight
(App.~\ref{sec:supp-bootstrap}), leaving no effect to trace across
weights. All fifteen runs share one machine and recipe (the
App.~\ref{sec:supp-training} recipe at effective batch 32 via gradient
accumulation, micro-batch 16, occlusion sub-batch scaled to preserve
the 4-per-32 exposure ratio, on an RTX~3070), so within-sweep
comparisons are like for like. Absolute levels are not comparable to
the main tables' models, which train on different hardware at true
batch 32.

Table~\ref{tab:supp-lambda} shows the grid. Four findings. First,
$\lambda$ behaves as a dial, not a cliff: noise drift falls
monotonically with $\lambda$ for every backbone (PointNet++ 0.250
$\rightarrow$ 0.009 across the grid), so the loss buys invariance
continuously. Second, $\lambda=0.5$ sits inside a broad operating
basin, not on a knife edge: for PointNet++ the neighboring grid points
bracket it benignly ($\lambda=0.25$ trades $+0.030$ clean accuracy for
$1.6\times$ the drift; $\lambda=1.0$ the reverse), and for the other
two backbones the $0.25$--$1.0$ span moves clean accuracy by at most
$0.057$. Where another value edges out $0.5$ on one metric we report
it as headroom and change nothing: retuning $\lambda$ on the
benchmark's own evaluation would be circular. Third, the collapse at
$\lambda=2.0$ is architecture-dependent, not universal: PointNet++
loses half its clean accuracy (0.473) with drift driven to zero, the
invariance term winning outright, while PointNet and RIConv train
normally at the same weight (0.900 and 0.893 clean). This is the same
species of training-stability failure as RISurConv's collapse at
$\lambda=0.5$ (Sec.~\ref{sec:generalization}), appearing at a
different point of the (architecture, $\lambda$) plane. Fourth,
RIConv's pose drift is at numerical zero at every $\lambda$
($|D| < 3\times10^{-4}$): architectural invariance is untouched by the
loss weight, exactly as construction promises.

One run in this grid required a disclosed repeat. The first
$\lambda=0.5$ PointNet++ run collapsed during training (validation
accuracy peaked at 0.700 by epoch 24 and never recovered, final clean
accuracy 0.680). A rerun under a fresh seed trained normally (0.820,
reported in Table~\ref{tab:supp-lambda}). Both runs ship in the
released CSVs. Seed replicates of the same configuration on the
campaign hardware (clean 0.910--0.930 across three seeds) put ordinary
seed noise near $\pm 0.01$, so the $0.14$ run-to-run gap is a sporadic
optimization failure of the kind this paper documents throughout
(RISurConv's collapse, the warmup arm's mixed effects,
$\lambda=2.0$), not $\lambda$-specific fragility: under the same
$\lambda$, three independent seeds trained without incident.

\begin{table*}[t]
\centering
\caption{Loss-weight ablation: held-out accuracy (clean, noise
$\sigma=0.05$) and noise drift $D$ at five values of $\lambda$, for
three backbones spanning both families, one run per point under the
recipe of App.~\ref{sec:supp-lambda}. Full 16-condition accuracies
and the occlusion and pose
drift series ship in the released CSVs.}
\label{tab:supp-lambda}
\footnotesize
\setlength{\tabcolsep}{3pt}
\begin{tabular}{lccccccccc}
\toprule
 & \multicolumn{3}{c}{PointNet++} & \multicolumn{3}{c}{PointNet} &
 \multicolumn{3}{c}{RIConv} \\
\cmidrule(lr){2-4}\cmidrule(lr){5-7}\cmidrule(lr){8-10}
$\lambda$ & clean$\uparrow$ & n.05$\uparrow$ & $D_n\downarrow$ & clean$\uparrow$ & n.05$\uparrow$ & $D_n\downarrow$ &
clean$\uparrow$ &
n.05$\uparrow$ &
$D_n\downarrow$ \\
\midrule
0.1  & 0.837 & 0.483 & 0.250 & 0.923 & 0.847 & 0.057 &
0.870 &
0.777 &
0.085 \\
0.25 & 0.850 & 0.430 & 0.140 & 0.937 & 0.797 & 0.036 &
0.890 &
0.790 &
0.045 \\
0.5  & 0.820 & 0.530 & 0.086 & 0.900 & 0.773 & 0.019 &
0.907 &
0.810 &
0.053 \\
1.0  & 0.787 & 0.417 & 0.052 & 0.900 & 0.797 & 0.020 &
0.850 &
0.733 &
0.047 \\
2.0  & 0.473 & 0.280 & 0.009 & 0.900 & 0.817 & 0.011 &
0.893 &
0.737 &
0.029 \\
\bottomrule
\end{tabular}
\end{table*}

\section{Seed Replication Study}
\label{sec:supp-seeds}

Every ModelNet40 model in this paper was retrained from scratch under
three fresh training seeds (10 architectures $\times$ \{baseline,
invariance loss\} $\times$ 3 seeds, 60 models, one unified recipe,
App.~\ref{sec:supp-training}), and each replicate ran the full
evaluation battery: the five drift axes, the 16-condition accuracy
evaluation, ModelNet40-C, and the head-independent probes.
Table~\ref{tab:supp-seeds} reports the ModelNet40-C headline metrics
as mean $\pm$ sd over seeds. Per-replicate values, drift statistics,
and probe accuracies for all 300 (model, axis, replicate) drift cells
ship with the artifact. One exception to the three fresh seeds:
replicate 3 of invariance-trained VN-PointNet reused the original
training run's checkpoint instead of training a new one (the campaign
skips any model whose checkpoint already exists), so that model's
statistics cover two fresh seeds plus the original run and its
standard deviations may be understated.

Four findings. First, the intervention's effects are seed-stable
where they are large: PointNet, PointNet++, PointMLP, and RIConv++
gain corrupted accuracy significantly in every one of the three
replicates (per-replicate paired bootstrap,
App.~\ref{sec:supp-bootstrap}). RISurConv regresses significantly in
every replicate, and the remaining models hover near zero with
seed-dependent sign. Second, single-model rankings are \emph{not}
seed-stable, exactly as the power analysis's scope disclaimer
anticipated: the best corrupted accuracy belongs to invariance-trained
PointMLP in replicate 1 and invariance-trained PointNet++ in
replicates 2 and 3. What is stable is the top tier's composition:
invariance-trained RIConv++ is in the top two in all three replicates
and is the only model whose mCE is below the DGCNN reference in every
seed ($0.935 \pm 0.020$). This makes it the most reproducible top
performer in the study, on par with invariance-trained
PointNet++ on corrupted accuracy ($0.821 \pm 0.009$ against
$0.823 \pm 0.020$). Third, RISurConv's training collapse is a
property of the configuration, not of one unlucky seed: its
invariance-loss arm collapses in all three replicates (clean accuracy
$0.724 \pm 0.052$, mCE rank 19--20 of 20 in every replicate). This
upgrades Sec.~\ref{sec:generalization}'s schedule-sensitivity
diagnosis from an observation to a reproducible phenomenon. Fourth,
the equivariant family's position is seed-stable: the two VN backbones
occupy the bottom tier on corruption robustness in every replicate,
under both training conditions.

\begin{table}[t]
\centering
\caption{Seed replication study: ModelNet40-C clean accuracy, mean
corrupted accuracy, and mCE as mean $\pm$ sd over three fresh training
seeds per model (each replicate's mCE normalized to its own DGCNN
baseline). The VN backbones are equivariant, not rotation-invariant.
$^\dagger$Two fresh seeds plus the original training run
(App.~\ref{sec:supp-seeds}).}
\label{tab:supp-seeds}
\scriptsize
\setlength{\tabcolsep}{2.5pt}
\begin{tabular}{llccc}
\toprule
Arch & Loss & Clean$\uparrow$ & Corr.$\uparrow$ & mCE$\downarrow$ \\
\midrule
PointNet & no & $0.928 \pm 0.026$ & $0.683 \pm 0.034$ & $1.887 \pm 0.340$ \\
PointNet++ & no & $0.903 \pm 0.028$ & $0.773 \pm 0.031$ & $1.464 \pm 0.278$ \\
DGCNN & no & $0.937 \pm 0.003$ & $0.805 \pm 0.001$ & $1.000$ (ref.) \\
RIConv & no & $0.911 \pm 0.001$ & $0.753 \pm 0.019$ & $1.400 \pm 0.039$ \\
RIConv++ & no & $0.916 \pm 0.015$ & $0.790 \pm 0.014$ & $1.085 \pm 0.096$ \\
RISurConv & no & $0.896 \pm 0.004$ & $0.680 \pm 0.023$ & $1.923 \pm 0.192$ \\
PointMLP & no & $0.913 \pm 0.018$ & $0.732 \pm 0.039$ & $1.557 \pm 0.384$ \\
PointTransf. & no & $0.940 \pm 0.008$ & $0.781 \pm 0.003$ & $1.231 \pm 0.047$ \\
VN-PointNet & no & $0.882 \pm 0.005$ & $0.578 \pm 0.023$ & $2.720 \pm 0.286$ \\
VN-DGCNN & no & $0.880 \pm 0.008$ & $0.633 \pm 0.015$ & $2.334 \pm 0.210$ \\
\midrule
PointNet & yes & $0.926 \pm 0.017$ & $0.755 \pm 0.015$ & $1.412 \pm 0.196$ \\
PointNet++ & yes & $0.904 \pm 0.010$ & $\mathbf{0.823 \pm 0.020}$ & $1.111 \pm 0.100$ \\
DGCNN & yes & $0.941 \pm 0.003$ & $0.795 \pm 0.015$ & $1.144 \pm 0.093$ \\
RIConv & yes & $0.917 \pm 0.003$ & $0.798 \pm 0.025$ & $1.180 \pm 0.155$ \\
RIConv++ & yes & $0.934 \pm 0.007$ & $\mathbf{0.821 \pm 0.009}$ & $\mathbf{0.935 \pm 0.020}$ \\
RISurConv & yes & $0.724 \pm 0.052$ & $0.580 \pm 0.063$ & $2.973 \pm 0.465$ \\
PointMLP & yes & $0.932 \pm 0.019$ & $0.798 \pm 0.023$ & $1.141 \pm 0.188$ \\
PointTransf. & yes & $0.934 \pm 0.001$ & $0.784 \pm 0.011$ & $1.216 \pm 0.102$ \\
VN-PointNet$^\dagger$ & yes & $0.888 \pm 0.013$ & $0.607 \pm 0.005$ & $2.530 \pm 0.100$ \\
VN-DGCNN & yes & $0.891 \pm 0.012$ & $0.603 \pm 0.020$ & $2.553 \pm 0.056$ \\
\bottomrule
\end{tabular}
\end{table}

\section{Disjoint Class-Subset Check}
\label{sec:supp-w5}

All ModelNet40 comparisons in this paper use one fixed 10-class
subset (Sec.~\ref{sec:design}). As a sensitivity check on that choice, we retrain
the two headline-comparison architectures, PointNet++ and RIConv++,
under both training conditions on a second, disjoint, non-degenerate
10-class subset with the same split sizes and recipe, and evaluate
them on ModelNet40-C through a subset-B class map
(Table~\ref{tab:supp-w5}). The comparison's shape reproduces. The
loss lifts corrupted accuracy for both architectures ($+0.023$ for
PointNet++, $+0.051$ for RIConv++), invariance-trained RIConv++ leads
the pair, and PointNet++ pays clean accuracy for its gain. Absolute
levels are lower throughout because the subset-B classes are harder,
which is why we compare shapes rather than values. This is a
single-seed sensitivity check, not a replication study, and it shows
the headline two-family comparison is not an artifact of the class
subset.

\begin{table}[t]
\centering
\caption{Disjoint class-subset check: the two headline-comparison
architectures retrained from scratch on a second 10-class ModelNet40
subset (disjoint from the paper's, same split sizes and recipe, one
seed) and evaluated on ModelNet40-C with a subset-B class map. mCE is
omitted because its reference is not comparable across subsets.}
\label{tab:supp-w5}
\small
\begin{tabular}{llcc}
\toprule
Arch & Loss & Clean$\uparrow$ & Corr.$\uparrow$ \\
\midrule
PointNet++ & no & 0.717 & 0.588 \\
PointNet++ & yes & 0.672 & 0.611 \\
RIConv++ & no & 0.819 & 0.700 \\
RIConv++ & yes & 0.843 & 0.751 \\
\bottomrule
\end{tabular}
\end{table}

\section{Rotation Sweep Per-Angle Structure}
\label{sec:supp-crossover}

Table~\ref{tab:rotation_gap} pools the deficit against RIConv++ over
all 144 (axis, angle) cells per backbone. This appendix reports the
per-angle structure behind that pooling.
Table~\ref{tab:supp-yaw-gap} gives the full per-angle gap for
PointNet++, the paper's overall winner, on yaw: all 36 angles lie
between 0.073 and 0.127, the gap is 0.103 at zero rotation, and the
other three axes behave identically (per-axis means 0.096 to 0.113).
At $n=300$ the per-cell standard error is 0.015 to 0.024 over the
observed accuracy range of 0.69 to 0.95, so this flatness is
resolved, not assumed.

An earlier $n=30$ version of this sweep appeared to show
angle-localized crossovers, with the deficit first sustaining at 10
to 110 degrees depending on axis and never sustaining on yaw. Those
angles were artifacts of 1/30 accuracy quantization (per-cell
standard error 0.03 to 0.08) and do not survive the tenfold larger
sample: the deficit is present at every angle including the
canonical pose, on every axis. We report this correction explicitly
because crossover angles are the natural quantity a reader would
extract from Figure~\ref{fig:rotation_sweep}, and at descriptive
sample sizes they are exactly the kind of number sampling noise
invents.

\begin{table}[h]
\centering
\caption{Full per-angle accuracy gap (RIConv++ minus the
widened-augmentation PointNet++) on the yaw axis, $n=300$. Every
angle sits within $[0.073, 0.127]$: the deficit neither closes nor
spikes anywhere. The other three axes are equally flat.}
\label{tab:supp-yaw-gap}
\small
\begin{tabular}{cc|cc|cc}
\toprule
Angle & Gap$\downarrow$ & Angle & Gap$\downarrow$ & Angle & Gap$\downarrow$ \\
\midrule
0   & 0.103 & 120 & 0.113 & 240 & 0.107 \\
10  & 0.103 & 130 & 0.100 & 250 & 0.113 \\
20  & 0.097 & 140 & 0.110 & 260 & 0.107 \\
30  & 0.083 & 150 & 0.087 & 270 & 0.107 \\
40  & 0.113 & 160 & 0.093 & 280 & 0.073 \\
50  & 0.103 & 170 & 0.100 & 290 & 0.090 \\
60  & 0.107 & 180 & 0.113 & 300 & 0.113 \\
70  & 0.083 & 190 & 0.087 & 310 & 0.097 \\
80  & 0.127 & 200 & 0.097 & 320 & 0.087 \\
90  & 0.113 & 210 & 0.097 & 330 & 0.097 \\
100 & 0.110 & 220 & 0.123 & 340 & 0.087 \\
110 & 0.113 & 230 & 0.100 & 350 & 0.110 \\
\bottomrule
\end{tabular}
\end{table}

\paragraph{Flatness is interpolation, not memorization.} The widened
augmentation itself is coarser than this sweep: training samples
discrete $30^{\circ}$-spaced steps (wrapping to all twelve step
positions around the circle, since $\pm270^{\circ}$ exceeds a
half-turn) with only $\pm8^{\circ}$ of jitter per step
(App.~\ref{sec:supp-training}), leaving roughly $14^{\circ}$-wide gaps
the model never samples during training. The $10^{\circ}$-resolution
sweep tests both regimes. Splitting Table~\ref{tab:supp-yaw-gap}'s 36
angles accordingly gives a mean gap of 0.103 at the 12 trained-step
angles against 0.101 at the 24 untrained gap angles, indistinguishable
at the 0.015--0.024 per-cell standard error. The flat offset in
Fig.~\ref{fig:rotation_sweep} is therefore genuine interpolation
across untrained angles, not an artifact of the training grid.

\section{Severity Test on the Equivariant Cells}
\label{sec:supp-vnseverity}

The secondary severity test was originally run on the 8-architecture
grid. We extend it to the 50 VN cells with the identical construction:
a randomized-block one-way ANOVA over the 5 severity levels with
instances as blocks, and the Friedman omnibus alongside. All 10 noise
cells and all 10 occlusion cells reject at $p<10^{-4}$, with $F$ from
328.8 to 1217.6 on noise and 26.3 to 254.4 on occlusion, above the
8-architecture grid's floors, so Sec.~\ref{sec:tradeoff}'s severity claims hold at the
full 10-backbone scope. All 30 pose cells fail to reject under the
ANOVA ($F \leq 2.2$), which restates the VN models' built-in
invariance exactly as RIConv's flat cells do, and one cell
(VN-PointNet, ABO roll) is byte-identical across levels. In 4 of the
30 pose cells the rank-based Friedman cross-check alone reaches
$p<0.05$. The response ranges in those cells are at most
$3.6\times10^{-4}$ in cosine similarity, float-level variation with no
structure any accuracy metric could see, and we read them as rank
artifacts of a numerically constant response rather than severity
effects. The per-cell statistics ship with the artifact.

\section{Per-Cell Confidence Intervals}
\label{sec:supp-ci}

Tables~\ref{tab:supp-pose}--\ref{tab:supp-occlusion} report the full
95\% confidence interval (mean $D$ with $[\text{CI}_{\text{low}},
\text{CI}_{\text{high}}]$, $\alpha=0.05$, same one-sided paired $t$
construction described in Sec.~\ref{sec:design}) for every one of the 250 primary
cells this paper's main text and Fig.~\ref{fig:teaser} draw from: pose
(yaw, tilt, and roll axes), noise, and occlusion (random camera), each
10 backbones (8 architectures plus the equivariant VN-PointNet and
VN-DGCNN) $\times$ 5 domains. ModelNet40 cells are replicate 1 of the seed study
(App.~\ref{sec:supp-seeds}), and the other domains are the original
runs. These are the same primary
confirmatory tests summarized as point estimates in the main paper
(Table~\ref{tab:noise} for noise). The main text's pose and occlusion
claims in Sec.~\ref{sec:tradeoff} are drawn from these same cells, and
Fig.~\ref{fig:teaser}'s pose axis is the mean of the 15 per-domain
yaw, tilt, and roll cells in
Tables~\ref{tab:supp-pose}--\ref{tab:supp-pose-roll}. The
$^{\text{ns}}$ marks inherit the one-sided construction: the primary
test is aimed at $H_0\colon D \leq 0$ and cannot flag drift in the
negative direction. Roll exposes this: on ModelNet40, PointNet
($-0.210$ $[-0.238, -0.182]$) and DGCNN ($-0.021$ $[-0.034, -0.007]$)
drift significantly \emph{negative}, embeddings on average closer to
the clean reference after rotation, yet both cells are non-rejections
by construction in Table~\ref{tab:supp-pose-roll}. We flag this rather
than explain it (partial object symmetry under the roll sweep is a
candidate, untested). The practical consequence is sign cancellation
in Fig.~\ref{fig:teaser}'s pooled pose axis, which flatters these two
backbones' positions in particular. An unsigned $|D|$ pooling would
move the standard backbones right while leaving the RI family's
near-zero cells unchanged, so the family-level trade-off is, if
anything, understated. The per-architecture pose positions of
PointNet and DGCNN should be read with this caveat.

\begin{table*}[t]
\centering
\caption{Per-cell 95\% confidence intervals for pose (yaw) embedding drift $D$: mean [CI low, CI high], all $n$ between 217 and 740. The ModelNet40 column is replicate 1 (App.~\ref{sec:supp-seeds}), and the other domains are the original study runs. VN-PointNet and VN-DGCNN are equivariant, not rotation-invariant. $^{\text{ns}}$ marks the 24 of 50 cells (spanning PointNet++, RIConv, RIConv++, RISurConv, PointMLP, and both VN models) where the primary one-sided paired $t$ test fails to reject $H_0$ at $\alpha=0.05$.}
\label{tab:supp-pose}
\footnotesize
\setlength{\tabcolsep}{3pt}
\begin{tabular}{lccccc}
\toprule
$D\downarrow$ & M40 & ShapeNet & ABO & ScanObj & S3DIS \\
\midrule
PointNet & 0.259 [0.235, 0.283] & 0.056 [0.046, 0.067] & 0.015 [0.009, 0.021] & 0.046 [0.029, 0.063] & 0.057 [0.034, 0.080] \\
PointNet++ & 0.066 [0.058, 0.075] & 0.019 [0.016, 0.023] & 0.001 [-0.002, 0.005]$^{\text{ns}}$ & 0.018 [0.012, 0.025] & 0.029 [0.018, 0.039] \\
DGCNN & 0.169 [0.152, 0.185] & 0.011 [0.009, 0.014] & 0.018 [0.011, 0.025] & 0.017 [0.011, 0.022] & 0.035 [0.025, 0.045] \\
RIConv & -0.000 [-0.001, 0.000]$^{\text{ns}}$ & 0.000 [-0.000, 0.000]$^{\text{ns}}$ & -0.000 [-0.001, 0.000]$^{\text{ns}}$ & 0.001 [-0.001, 0.002]$^{\text{ns}}$ & 0.000 [-0.001, 0.001]$^{\text{ns}}$ \\
RIConv++ & -0.000 [-0.002, 0.001]$^{\text{ns}}$ & 0.003 [0.000, 0.005] & 0.000 [-0.001, 0.001]$^{\text{ns}}$ & 0.004 [-0.001, 0.008] & 0.001 [-0.002, 0.004]$^{\text{ns}}$ \\
RISurConv & 0.002 [0.000, 0.004] & 0.002 [-0.001, 0.005]$^{\text{ns}}$ & 0.004 [0.001, 0.007] & 0.007 [-0.004, 0.019]$^{\text{ns}}$ & 0.008 [-0.006, 0.021]$^{\text{ns}}$ \\
PointMLP & 0.023 [0.019, 0.028] & 0.005 [0.003, 0.008] & 0.000 [-0.002, 0.002]$^{\text{ns}}$ & -0.001 [-0.004, 0.001]$^{\text{ns}}$ & 0.010 [0.006, 0.014] \\
PointTransf. & 0.092 [0.080, 0.104] & 0.016 [0.011, 0.020] & 0.004 [0.002, 0.005] & 0.010 [0.006, 0.015] & 0.017 [0.010, 0.025] \\
VN-PointNet & 0.000 [-0.000, 0.000]$^{\text{ns}}$ & 0.000 [-0.000, 0.000]$^{\text{ns}}$ & 0.000 [0.000, 0.000]$^{\text{ns}}$ & -0.000 [-0.000, 0.000]$^{\text{ns}}$ & 0.000 [-0.000, 0.000]$^{\text{ns}}$ \\
VN-DGCNN & -0.000 [-0.000, 0.000]$^{\text{ns}}$ & -0.000 [-0.000, 0.000]$^{\text{ns}}$ & -0.000 [-0.000, 0.000]$^{\text{ns}}$ & -0.000 [-0.000, 0.000]$^{\text{ns}}$ & 0.000 [-0.000, 0.000]$^{\text{ns}}$ \\
\bottomrule
\end{tabular}
\end{table*}

\begin{table*}[t]
\centering
\caption{Per-cell 95\% confidence intervals for pose (tilt) embedding drift $D$: mean [CI low, CI high], all $n$ between 217 and 740. The ModelNet40 column is replicate 1 (App.~\ref{sec:supp-seeds}), and the other domains are the original study runs. VN-PointNet and VN-DGCNN are equivariant, not rotation-invariant. $^{\text{ns}}$ marks the 23 of 50 cells (spanning every backbone except DGCNN) where the primary one-sided paired $t$ test fails to reject $H_0$ at $\alpha=0.05$.}
\label{tab:supp-pose-tilt}
\footnotesize
\setlength{\tabcolsep}{3pt}
\begin{tabular}{lccccc}
\toprule
$D\downarrow$ & M40 & ShapeNet & ABO & ScanObj & S3DIS \\
\midrule
PointNet & 0.262 [0.236, 0.288] & 0.117 [0.081, 0.152] & 0.251 [0.192, 0.310] & 0.069 [0.030, 0.108] & 0.024 [-0.023, 0.071]$^{\text{ns}}$ \\
PointNet++ & 0.024 [0.009, 0.039] & 0.121 [0.097, 0.144] & 0.145 [0.114, 0.175] & 0.073 [0.051, 0.095] & -0.016 [-0.029, -0.002]$^{\text{ns}}$ \\
DGCNN & 0.192 [0.170, 0.213] & 0.188 [0.168, 0.208] & 0.261 [0.222, 0.300] & 0.098 [0.073, 0.124] & 0.028 [0.006, 0.049] \\
RIConv & -0.000 [-0.001, 0.000]$^{\text{ns}}$ & 0.000 [-0.000, 0.000]$^{\text{ns}}$ & -0.000 [-0.001, 0.000]$^{\text{ns}}$ & 0.001 [-0.001, 0.002]$^{\text{ns}}$ & 0.000 [-0.001, 0.001]$^{\text{ns}}$ \\
RIConv++ & 0.004 [0.002, 0.006] & 0.002 [0.001, 0.003] & 0.001 [-0.001, 0.002]$^{\text{ns}}$ & 0.003 [0.001, 0.006] & 0.002 [-0.000, 0.003] \\
RISurConv & 0.002 [-0.000, 0.004] & -0.004 [-0.006, -0.001]$^{\text{ns}}$ & 0.004 [-0.002, 0.009]$^{\text{ns}}$ & 0.003 [-0.007, 0.013]$^{\text{ns}}$ & -0.004 [-0.018, 0.011]$^{\text{ns}}$ \\
PointMLP & 0.045 [0.038, 0.051] & 0.264 [0.239, 0.288] & 0.277 [0.234, 0.321] & 0.156 [0.127, 0.185] & -0.028 [-0.042, -0.015]$^{\text{ns}}$ \\
PointTransf. & 0.172 [0.153, 0.190] & 0.300 [0.268, 0.331] & 0.342 [0.301, 0.383] & 0.115 [0.084, 0.146] & -0.018 [-0.034, -0.002]$^{\text{ns}}$ \\
VN-PointNet & 0.000 [-0.000, 0.000]$^{\text{ns}}$ & 0.000 [-0.000, 0.000] & 0.000 [-0.000, 0.000]$^{\text{ns}}$ & -0.000 [-0.000, 0.000]$^{\text{ns}}$ & -0.000 [-0.000, 0.000]$^{\text{ns}}$ \\
VN-DGCNN & -0.000 [-0.000, 0.000]$^{\text{ns}}$ & 0.000 [-0.000, 0.000]$^{\text{ns}}$ & -0.000 [-0.000, 0.000]$^{\text{ns}}$ & -0.000 [-0.000, 0.000]$^{\text{ns}}$ & -0.000 [-0.000, 0.000]$^{\text{ns}}$ \\
\bottomrule
\end{tabular}
\end{table*}

\begin{table*}[t]
\centering
\caption{Per-cell 95\% confidence intervals for pose (roll) embedding drift $D$: mean [CI low, CI high], all $n$ between 217 and 740. The ModelNet40 column is replicate 1 (App.~\ref{sec:supp-seeds}), and the other domains are the original study runs. VN-PointNet and VN-DGCNN are equivariant, not rotation-invariant. $^{\text{ns}}$ marks the 32 of 50 cells (spanning all 10 backbones) where the primary one-sided paired $t$ test fails to reject $H_0$ at $\alpha=0.05$. Several are significantly \emph{negative} (see App.~\ref{sec:supp-ci} text).}
\label{tab:supp-pose-roll}
\footnotesize
\setlength{\tabcolsep}{3pt}
\begin{tabular}{lccccc}
\toprule
$D\downarrow$ & M40 & ShapeNet & ABO & ScanObj & S3DIS \\
\midrule
PointNet & -0.210 [-0.238, -0.182]$^{\text{ns}}$ & -0.045 [-0.071, -0.019]$^{\text{ns}}$ & 0.097 [0.050, 0.144] & 0.028 [0.009, 0.048] & -0.012 [-0.039, 0.016]$^{\text{ns}}$ \\
PointNet++ & 0.005 [-0.004, 0.013]$^{\text{ns}}$ & -0.015 [-0.038, 0.009]$^{\text{ns}}$ & 0.106 [0.083, 0.129] & 0.039 [0.019, 0.059] & -0.009 [-0.022, 0.005]$^{\text{ns}}$ \\
DGCNN & -0.021 [-0.034, -0.007]$^{\text{ns}}$ & 0.116 [0.097, 0.136] & 0.154 [0.122, 0.186] & 0.058 [0.035, 0.081] & -0.032 [-0.047, -0.018]$^{\text{ns}}$ \\
RIConv & -0.000 [-0.001, 0.000]$^{\text{ns}}$ & 0.000 [-0.000, 0.000]$^{\text{ns}}$ & -0.000 [-0.001, 0.000]$^{\text{ns}}$ & 0.000 [-0.001, 0.001]$^{\text{ns}}$ & 0.000 [-0.001, 0.001]$^{\text{ns}}$ \\
RIConv++ & -0.002 [-0.003, -0.000]$^{\text{ns}}$ & -0.002 [-0.005, 0.001]$^{\text{ns}}$ & -0.003 [-0.009, 0.003]$^{\text{ns}}$ & -0.001 [-0.005, 0.004]$^{\text{ns}}$ & 0.002 [-0.001, 0.005]$^{\text{ns}}$ \\
RISurConv & 0.002 [0.000, 0.004] & 0.005 [0.002, 0.009] & 0.026 [0.016, 0.036] & 0.006 [-0.004, 0.015]$^{\text{ns}}$ & 0.010 [-0.001, 0.021] \\
PointMLP & 0.008 [0.003, 0.013] & 0.189 [0.159, 0.218] & 0.290 [0.246, 0.334] & 0.092 [0.068, 0.115] & -0.024 [-0.036, -0.013]$^{\text{ns}}$ \\
PointTransf. & -0.012 [-0.025, 0.000]$^{\text{ns}}$ & 0.206 [0.176, 0.236] & 0.274 [0.230, 0.317] & 0.063 [0.031, 0.095] & -0.019 [-0.029, -0.010]$^{\text{ns}}$ \\
VN-PointNet & -0.000 [-0.000, 0.000]$^{\text{ns}}$ & -0.000 [-0.000, 0.000]$^{\text{ns}}$ & 0.000 [0.000, 0.000]$^{\text{ns}}$ & 0.000 [-0.000, 0.000]$^{\text{ns}}$ & 0.000 [-0.000, 0.000]$^{\text{ns}}$ \\
VN-DGCNN & -0.000 [-0.000, 0.000]$^{\text{ns}}$ & 0.000 [-0.000, 0.000]$^{\text{ns}}$ & 0.000 [-0.000, 0.000]$^{\text{ns}}$ & -0.000 [-0.000, 0.000]$^{\text{ns}}$ & -0.000 [-0.000, 0.000]$^{\text{ns}}$ \\
\bottomrule
\end{tabular}
\end{table*}

\begin{table*}[t]
\centering
\caption{Per-cell 95\% confidence intervals for noise embedding drift $D$: mean [CI low, CI high], all $n$ between 217 and 740 (main-paper Table~\ref{tab:noise} reports the point estimates only). The ModelNet40 column is replicate 1 (App.~\ref{sec:supp-seeds}), and the other domains are the original study runs. VN-PointNet and VN-DGCNN are equivariant, not rotation-invariant. All 50 cells reject $H_0$ at $\alpha=0.05$.}
\label{tab:supp-noise}
\footnotesize
\setlength{\tabcolsep}{3pt}
\begin{tabular}{lccccc}
\toprule
$D\downarrow$ & M40 & ShapeNet & ABO & ScanObj & S3DIS \\
\midrule
PointNet & 0.143 [0.130, 0.157] & 0.170 [0.152, 0.187] & 0.268 [0.231, 0.304] & 0.485 [0.437, 0.532] & 0.763 [0.691, 0.835] \\
PointNet++ & 0.454 [0.429, 0.480] & 0.333 [0.307, 0.358] & 0.712 [0.665, 0.758] & 0.844 [0.802, 0.885] & 0.880 [0.817, 0.942] \\
DGCNN & 0.105 [0.098, 0.112] & 0.250 [0.228, 0.272] & 0.230 [0.204, 0.256] & 0.367 [0.338, 0.397] & 0.583 [0.531, 0.636] \\
RIConv & 0.447 [0.424, 0.469] & 0.582 [0.548, 0.617] & 0.647 [0.598, 0.696] & 0.712 [0.665, 0.760] & 0.655 [0.598, 0.713] \\
RIConv++ & 0.537 [0.512, 0.561] & 0.561 [0.529, 0.592] & 0.673 [0.630, 0.716] & 0.774 [0.727, 0.821] & 0.752 [0.691, 0.814] \\
RISurConv & 0.529 [0.499, 0.559] & 0.671 [0.633, 0.710] & 0.764 [0.718, 0.811] & 0.744 [0.691, 0.797] & 0.836 [0.760, 0.912] \\
PointMLP & 0.264 [0.245, 0.282] & 0.084 [0.072, 0.095] & 0.331 [0.294, 0.367] & 0.388 [0.348, 0.427] & 0.359 [0.308, 0.411] \\
PointTransf. & 0.149 [0.132, 0.165] & 0.122 [0.106, 0.138] & 0.224 [0.192, 0.257] & 0.428 [0.383, 0.472] & 0.659 [0.589, 0.728] \\
VN-PointNet & 0.332 [0.315, 0.349] & 0.362 [0.338, 0.385] & 0.495 [0.459, 0.531] & 0.576 [0.537, 0.615] & 0.619 [0.566, 0.671] \\
VN-DGCNN & 0.154 [0.143, 0.165] & 0.313 [0.288, 0.338] & 0.325 [0.294, 0.355] & 0.536 [0.496, 0.577] & 0.626 [0.572, 0.680] \\
\bottomrule
\end{tabular}
\end{table*}

\begin{table*}[t]
\centering
\caption{Per-cell 95\% confidence intervals for occlusion (random camera) embedding drift $D$: mean [CI low, CI high], all $n$ between 217 and 740. The ModelNet40 column is replicate 1 (App.~\ref{sec:supp-seeds}), and the other domains are the original study runs. VN-PointNet and VN-DGCNN are equivariant, not rotation-invariant. Note the absence of a consistent family-level pattern discussed in Sec.~\ref{sec:tradeoff}. All 50 cells reject $H_0$ at $\alpha=0.05$.}
\label{tab:supp-occlusion}
\footnotesize
\setlength{\tabcolsep}{3pt}
\begin{tabular}{lccccc}
\toprule
$D\downarrow$ & M40 & ShapeNet & ABO & ScanObj & S3DIS \\
\midrule
PointNet & 0.273 [0.242, 0.304] & 0.203 [0.173, 0.234] & 0.470 [0.409, 0.531] & 0.571 [0.516, 0.627] & 0.272 [0.208, 0.336] \\
PointNet++ & 0.326 [0.298, 0.355] & 0.140 [0.118, 0.163] & 0.395 [0.343, 0.447] & 0.495 [0.444, 0.546] & 0.302 [0.239, 0.365] \\
DGCNN & 0.293 [0.267, 0.319] & 0.206 [0.180, 0.231] & 0.408 [0.357, 0.459] & 0.475 [0.427, 0.524] & 0.270 [0.219, 0.322] \\
RIConv & 0.373 [0.342, 0.403] & 0.236 [0.204, 0.267] & 0.434 [0.380, 0.488] & 0.506 [0.452, 0.559] & 0.293 [0.235, 0.351] \\
RIConv++ & 0.428 [0.397, 0.459] & 0.217 [0.188, 0.245] & 0.484 [0.436, 0.531] & 0.546 [0.498, 0.594] & 0.205 [0.150, 0.260] \\
RISurConv & 0.252 [0.214, 0.291] & 0.210 [0.177, 0.244] & 0.517 [0.463, 0.571] & 0.438 [0.381, 0.496] & 0.191 [0.129, 0.254] \\
PointMLP & 0.344 [0.315, 0.372] & 0.108 [0.078, 0.137] & 0.390 [0.338, 0.442] & 0.479 [0.429, 0.529] & 0.321 [0.260, 0.382] \\
PointTransf. & 0.336 [0.306, 0.366] & 0.162 [0.134, 0.190] & 0.440 [0.388, 0.492] & 0.496 [0.446, 0.545] & 0.274 [0.218, 0.331] \\
VN-PointNet & 0.284 [0.255, 0.313] & 0.163 [0.132, 0.194] & 0.429 [0.380, 0.478] & 0.502 [0.454, 0.550] & 0.180 [0.131, 0.230] \\
VN-DGCNN & 0.281 [0.255, 0.306] & 0.157 [0.127, 0.187] & 0.427 [0.375, 0.479] & 0.577 [0.528, 0.626] & 0.204 [0.153, 0.256] \\
\bottomrule
\end{tabular}
\end{table*}

\section{Pilot Calibration and Realized Power}
\label{sec:supp-pilot}

The per-domain required $n$ of Sec.~\ref{sec:design} was derived from deliberately
small pilots: one withheld instance per class (10 instances, 7 for
S3DIS), run on PointNet++ only, whose pilot $\mathrm{sd}_D$ fed the
noncentral-t calculation at the pre-registered $\delta=0.05$,
$\alpha=0.05$, 80\% target power. A fair question is whether a
variance estimate that thin, from a single architecture, can carry a
200-cell study. This section answers it against the realized data.

Scope: this calibration analysis covers the 8-architecture grid's 200
cells. The equivariant VN models' 50 cells (25 each for VN-PointNet
and VN-DGCNN), added later, use the same per-domain pools and
required-$n$ thresholds.

\paragraph{Calibration.} Across the 200 primary cells (ModelNet40 at
replicate 1, the other four domains from the original study), the
ratio of realized confirmatory $\mathrm{sd}_D$ to the corresponding
pilot estimate has median 0.93, so the typical pilot was nearly
unbiased.
The tails are real, though: ratios reach $11.5\times$ in cells where
PointNet++'s pilot variance was smallest, so individual pilots did
underestimate individual cells.

\paragraph{Consequence.} What decides whether that mattered is a
sensitivity reanalysis: recomputing required $n$ per cell under the
\emph{realized} $\mathrm{sd}_D$, same pre-registered
$\delta$/$\alpha$/power and the same noncentral-t formula. The
achieved confirmatory $n$ meets this stricter requirement in 150 of
200 cells, and the 50 shortfalls split into exactly two kinds. 49
rejected $H_0$ anyway, with observed mean $D$ from 0.097 to 0.880,
which is $2\times$ to $17.6\times$ the minimal effect the power
analysis was protecting. These cells were underpowered for detecting
$\delta$, but that is moot when the
true effect dwarfs $\delta$. The one remaining cell, PointNet on
S3DIS tilt (mean $D=0.024$, achieved power 0.68), is a genuinely
inconclusive non-rejection, marked \textsuperscript{ns} in
Table~\ref{tab:supp-pose-tilt}. No claim in this paper rests on it.
Conversely, every one of the other 49 non-rejections, including all
the RI-family flatness cells that anchor Sec.~\ref{sec:tradeoff}'s ``exact invariance
is real'' claim, carries achieved power of at least 0.97 (median
1.00) at $\delta$. This makes the accepted flat cells the best-powered
results in the study. The full 200-cell calibration table ships with
the artifact.

\paragraph{Occlusion severity ordering.} Sec.~\ref{sec:design}'s claim that farther
cameras remove more surface, verified empirically, quantifies as
follows on the pilot instances. Mean visible points (of 1024) fall
monotonically with the camera-distance multiplier in every domain,
e.g.\ 638 at multiplier 0.3 down to 308 at 6.0 on ModelNet40. The
severest level ranges from 298 (ABO) to 555 (S3DIS) visible
points across domains, and no domain shows a non-monotone step
anywhere in the ladder.

\section{Drift--Accuracy Commensurability}
\label{sec:supp-commensurability}

Cosine drift lives in each model's own embedding geometry, so a
skeptic can ask whether comparing mean $D$ \emph{across} architectures
(Table~\ref{tab:noise}'s orderings, Fig.~\ref{fig:teaser}, the
cross-architecture Friedman test) measures robustness or merely
embedding smoothness. We test this on ModelNet40, the one domain
where per-instance classification accuracy exists at every benchmark
severity (the 300 held-out queries of Sec.~\ref{sec:design}, 16 conditions each). We
recompute every cross-architecture comparison for the 8-architecture
grid under an accuracy
analog of $D$: accuracy at the mild reference severity minus accuracy
at the harshest, the exact severity pairs $D$ itself uses.
Accuracy is a decision-level quantity and is commensurable across
architectures by construction, so agreement between the two orderings
is evidence that the drift orderings track something decision-relevant
rather than geometry-specific.

Table~\ref{tab:supp-w2} shows the result, replicate 1 of the seed study
(matching Table~\ref{tab:noise}'s ModelNet40 column). Where the main
text makes cross-architecture drift claims, the two metrics agree
strongly: Spearman $\rho = 0.90$ on pose, $0.97$ on noise, and $0.98$
on occlusion, with $\rho = 0.95$ between pose drift and the
clean-minus-severe accuracy drop. The noise inversion that anchors
Sec.~\ref{sec:tradeoff} reproduces under accuracy: RIConv++ ($\Delta$acc $0.620$) and
RISurConv ($0.497$) are the two most fragile of all eight in both
metrics. Pose shows the widest rank disagreement of the three
perturbations ($\rho=0.90$, against $0.97$--$0.98$ for noise and
occlusion). The disagreement is driven by the three
architecturally-invariant RI
architectures' near-zero drift and near-zero accuracy delta
reordering among themselves ($D$ from $-0.0005$ to $0.002$,
$\Delta$acc from $-0.017$ to $-0.003$). Both metrics agree these
three are exactly invariant, disagreeing only on their internal
ordering at that scale. Occlusion, which the original single-seed
study corroborated only moderately ($\rho = 0.69$, the reason the main
text still declines to make an occlusion family-level claim in
Sec.~\ref{sec:tradeoff}), now agrees as strongly under replicate 1 as pose and noise.
The original caveat was conservative, not overturned. Pooled over all
24 (architecture, perturbation) cells, drift and its accuracy analog
correlate at Pearson $r = 0.96$.

Scope: this corroboration is ModelNet40-only, since the other four
domains have no per-severity accuracy evaluations. Cross-architecture
drift statements in those domains rest on the within-architecture
per-cell tests plus the calibration established here.

\begin{table}[t]
\centering
\caption{Drift--accuracy commensurability on ModelNet40, replicate 1 of
the seed study. For each
perturbation, mean embedding drift $D$ next to its accuracy analog
$\Delta\text{acc}$ (accuracy at the mild reference severity minus
accuracy at the harshest, same severity pairs as $D$. Pose is yaw,
occlusion the random camera).
Spearman rank agreement between the two columns: pose $\rho=0.90$,
noise $\rho=0.97$, occlusion $\rho=0.98$, and Pearson $r=0.96$ pooled over
all 24 cells.}
\label{tab:supp-w2}
\footnotesize
\setlength{\tabcolsep}{4pt}
\begin{tabular}{lcccccc}
\toprule
 & \multicolumn{2}{c}{Pose (yaw)} & \multicolumn{2}{c}{Noise} & \multicolumn{2}{c}{Occlusion} \\
 & $D\downarrow$ & $\Delta$acc$\downarrow$ & $D\downarrow$ & $\Delta$acc$\downarrow$ & $D\downarrow$ & $\Delta$acc$\downarrow$ \\
\midrule
PointNet & 0.259 & 0.240 & 0.143 & 0.160 & 0.273 & 0.213 \\
PointNet++ & 0.066 & 0.113 & 0.454 & 0.417 & 0.326 & 0.287 \\
DGCNN & 0.169 & 0.260 & 0.105 & 0.110 & 0.293 & 0.280 \\
RIConv & 0.000 & $-$0.003 & 0.447 & 0.440 & 0.373 & 0.350 \\
RIConv++ & 0.000 & $-$0.010 & 0.537 & 0.620 & 0.428 & 0.413 \\
RISurConv & 0.002 & $-$0.017 & 0.529 & 0.497 & 0.252 & 0.133 \\
PointMLP & 0.023 & 0.040 & 0.264 & 0.327 & 0.344 & 0.297 \\
PointTransf. & 0.092 & 0.127 & 0.149 & 0.160 & 0.336 & 0.323 \\
\bottomrule
\end{tabular}
\end{table}

\begin{table}[t]
\centering
\caption{Test-time PCA canonicalization on ModelNet40, replicate 1 of
the seed study (300 held-out
queries, nearest-class-mean head, references re-embedded under the same
arm). Each pair is standard $\rightarrow$ canonicalized input. The RI
architectures are unaffected, as architectural invariance
predicts. VN-PointNet/VN-DGCNN (equivariant) are likewise
unaffected on pose, consistent with their architectural pose-invariance, and
for the standard backbones canonicalization makes pose exactly
flat (canonicalized pose\,120 equals canonicalized clean within 0.020
everywhere) but pays for it in clean accuracy and noise.}
\label{tab:supp-canon}
\footnotesize
\setlength{\tabcolsep}{2.5pt}
\begin{tabular}{lcccc}
\toprule
 & Clean$\uparrow$ & Pose 120$\uparrow$ & Noise .05$\uparrow$ & Occl.\ 6.0$\uparrow$ \\
\midrule
PointNet & .90$\rightarrow$.58 & .61$\rightarrow$.58 & .76$\rightarrow$.51 & .48$\rightarrow$.31 \\
PointNet++ & .89$\rightarrow$.73 & .77$\rightarrow$.72 & .48$\rightarrow$.40 & .34$\rightarrow$.24 \\
DGCNN & .94$\rightarrow$.70 & .67$\rightarrow$.70 & .85$\rightarrow$.54 & .46$\rightarrow$.34 \\
RIConv & .91$\rightarrow$.90 & .90$\rightarrow$.90 & .44$\rightarrow$.43 & .31$\rightarrow$.31 \\
RIConv++ & .93$\rightarrow$.92 & .93$\rightarrow$.92 & .29$\rightarrow$.26 & .28$\rightarrow$.36 \\
RISurConv & .87$\rightarrow$.86 & .87$\rightarrow$.87 & .38$\rightarrow$.36 & .41$\rightarrow$.46 \\
PointMLP & .91$\rightarrow$.87 & .86$\rightarrow$.86 & .58$\rightarrow$.59 & .41$\rightarrow$.36 \\
PointTransf. & .92$\rightarrow$.80 & .80$\rightarrow$.81 & .74$\rightarrow$.61 & .44$\rightarrow$.38 \\
VN-PointNet & .85$\rightarrow$.86 & .85$\rightarrow$.86 & .50$\rightarrow$.49 & .39$\rightarrow$.32 \\
VN-DGCNN & .87$\rightarrow$.85 & .87$\rightarrow$.85 & .72$\rightarrow$.72 & .32$\rightarrow$.31 \\
\bottomrule
\end{tabular}
\end{table}

\section{Test-Time Canonicalization Baseline}
\label{sec:supp-canon}

The cheapest classical route to pose robustness is neither an RI
architecture nor a training objective: rotate each observed cloud into
its own PCA frame at test time and use a standard backbone. This
section prices that route on the paper's own protocol. Each of the 300
held-out ModelNet40 queries is perturbed (the 16 conditions of Sec.~\ref{sec:design}'s
accuracy evaluation), then canonicalized: axes ordered by descending
covariance eigenvalue, the two leading signs fixed by positive third
moment, the third by right-handedness. Both arms, standard and
canonicalized, share the same perturbed cloud and the same
per-replicate sample indices, so every comparison is paired. Because
the PCA frame is not the dataset's canonical pose, canonicalized
queries are scored against gallery references re-embedded under the
same canonicalization, using the training-free heads of Sec.~\ref{sec:design} on the
frozen checkpoints (the trained heads, whose inputs canonicalization
moves out of distribution by construction, are included in the
released CSVs but not read as the fair comparison).

Table~\ref{tab:supp-canon} shows the result, replicate 1 of the seed
study, for the 8 architectures plus the two equivariant VN baselines.
Three findings. First, canonicalization delivers what it promises:
canonicalized pose accuracy equals canonicalized clean accuracy within
0.020 at every angle for every architecture, exact invariance by
construction, and the RI family is untouched by the procedure (clean
$0.903 \rightarrow 0.891$ family mean), which doubles as a pipeline
sanity check. VN-PointNet and VN-DGCNN are likewise untouched on
pose (clean and pose-120 already match under the standard arm alone,
before any canonicalization), consistent with their architectural
equivariance. Canonicalizing input into an already-equivariant
backbone is redundant, not harmful, but it buys nothing either. Second, the price
is severe: the standard backbones'
family-mean clean accuracy falls from 0.913 to 0.737, with the cost
strongly architecture-dependent (PointNet $-0.32$, PointMLP $-0.04$).
Third, and most diagnostic: canonicalization \emph{imports} the RI
family's noise fragility. Family-mean accuracy at $\sigma=0.05$ drops
from 0.683 to 0.527, approaching the RI family's own 0.369, because
the covariance eigenvectors that define the frame are themselves
perturbation-sensitive. This is the global analog of the
noise-sensitive
local reference axes Sec.~\ref{sec:tradeoff} identifies in the RI
family. The noise tax follows the frame estimator, wherever the frame
is computed. Scope: this is test-time canonicalization on frozen
backbones. Retraining on canonicalized inputs is the stronger variant
and is not run here.

\section{Head-Independent Probe Details}
\label{sec:supp-probe}

\begin{table}[t]
\centering
\caption{Invariance-loss accuracy delta (points) on the two headline
conditions, under the trained head and the three training-free probes,
all 10 architectures (replicate 1). Occlusion gains are positive under
every head for every architecture except collapsed RISurConv. Noise
gains are positive everywhere except DGCNN (three of four heads) and
VN-DGCNN's subspace head, with RIConv++ gaining most of all.}
\label{tab:probe-delta}
\scriptsize
\setlength{\tabcolsep}{3pt}
\begin{tabular}{lcccccccc}
\toprule
 & \multicolumn{4}{c}{Occlusion (mult.\ 6.0)$\uparrow$} & \multicolumn{4}{c}{Noise ($\sigma=0.05$)$\uparrow$} \\
\cmidrule(lr){2-5}\cmidrule(lr){6-9}
 & head & sm & knn & ncm & head & sm & knn & ncm \\
\midrule
PointNet & +19.7 & +16.3 & +17.3 & +19.0 & +12.3 & +5.3 & +7.0 & +11.0 \\
PointNet++ & +18.3 & +25.0 & +15.7 & +16.7 & +22.3 & +1.7 & +14.7 & +21.0 \\
DGCNN & +24.7 & +16.3 & +18.0 & +22.7 & $-$3.3 & $-$2.0 & +2.7 & $-$1.7 \\
PointMLP & +35.0 & +32.7 & +33.7 & +35.7 & +27.0 & +16.0 & +24.3 & +27.7 \\
PointTransf. & +24.3 & +20.7 & +22.3 & +24.7 & +12.0 & +13.0 & +13.0 & +12.0 \\
RIConv & +23.3 & +15.0 & +19.3 & +23.0 & +38.7 & +34.3 & +40.7 & +41.3 \\
RIConv++ & \textbf{+46.3} & +32.0 & +40.3 & \textbf{+45.0} & \textbf{+52.7} & \textbf{+45.0} & \textbf{+52.3} & \textbf{+52.7} \\
RISurConv & $-$6.3 & $-$9.0 & $-$2.7 & $-$4.3 & +6.7 & +0.3 & +0.3 & +5.3 \\
VN-PointNet & +19.0 & +12.3 & +19.0 & +21.0 & +21.7 & +12.0 & +19.7 & +25.7 \\
VN-DGCNN & +29.7 & +19.3 & +25.7 & +28.0 & +2.0 & $-$4.7 & +0.3 & +6.3 \\
\bottomrule
\end{tabular}
\end{table}

Protocol (ModelNet40). Reference vectors are extracted from the
450-instance gallery split, 45 instances per class with all 5 sampling
replicates kept as separate vectors, giving 225 references per class.
Queries are the 300-instance held-out test split under the 16
conditions used elsewhere in the paper (clean, 5 yaw angles, 5 noise
levels, 5 occlusion levels). Each query averages the same 5 replicates
and is re-normalized, so all heads see inputs identical to the trained
head's. The pre-normalization embedding for the Euclidean control is
captured with a forward hook on the shared embedding-head module
every backbone terminates with. The trained-head accuracies recorded
in the same pass reproduce the paper's own tables (exactly on
deterministic conditions, within $\sim$2 points on stochastic ones),
which is the evidence the extraction protocol is faithful.
Tables~\ref{tab:supp-probe} and~\ref{tab:supp-probe2} report all heads
on the four headline conditions for all 20 models (replicate-1
checkpoints throughout this appendix, App.~\ref{sec:supp-seeds}).

\paragraph{Subspace-dimension sensitivity.} On PointNet, clean probe
accuracy is 0.927 at subspace dimension 1, dips to 0.860 at 5, peaks
at 0.940 for 10--15, and is flat at 0.933 from 20 through 40, so the
choice of dimension 20 is not load-bearing. (This sweep ran on the
original-study checkpoint. The dimension choice, not the accuracy
level, is what it establishes.)

\paragraph{Euclidean control.} Occlusion deltas
(invariance-trained minus baseline) under the Euclidean
nearest-class-mean on pre-normalization features: PointNet $+1.0$,
PointNet++ $+1.7$, DGCNN $+10.0$, PointMLP $+14.7$, PointTransformer
$+10.7$. Each is a fraction of the same architecture's gain under the
normalized heads (PointNet $+16.3$ to $+19.7$, PointNet++ $+15.7$ to
$+25.0$). The invariance loss constrains embedding direction, not
magnitude, so a magnitude-sensitive head is not expected to see the
angular-geometry gains fully (Sec.~\ref{sec:generalization}). RIConv
is the sharpest case: its invariance-trained model reads at 0.800
Euclidean clean accuracy against 0.867 for its baseline and
0.91--0.93 under every normalized-space head, and its Euclidean noise
gain ($+21.7$) is roughly half its normalized-head gains ($+34.3$ to
$+41.3$).

\paragraph{Subspace-head weakness on occlusion.} For invariance-trained
models the subspace probe is the weakest (or tied-weakest) of the
three training-free heads on occlusion in 8 of the 10 architectures
(PointNet++ and PointTransformer are the exceptions), e.g.\ RIConv
0.430 against 0.503 (kNN) and 0.533 (class-mean). Occluded embeddings
appear to drift off the class subspace in a way projection length
penalizes more than a class-mean distance does, reinforcing the main
text's recommendation to report kNN or NCM alongside the subspace
method rather than in its place.

\paragraph{RISurConv's collapse, head by head.} Sec.~\ref{sec:generalization}
reports RISurConv's clean-accuracy collapse under the trained head
(0.892 to 0.753, replicate 1) and states that every probe head sees
the same drop. The numbers: head $0.873\!\to\!0.757$, subspace
$0.847\!\to\!0.643$, kNN $0.850\!\to\!0.763$, class-mean
$0.873\!\to\!0.763$, Euclidean $0.873\!\to\!0.760$. No re-heading of the
frozen embedding recovers any of it, which is what licenses calling
this representation damage rather than a classifier artifact.
Consistent with this instability, the sign of RISurConv's occlusion
probe delta also varies across the three training seeds (negative in
replicate 1, positive in replicates 2 and 3), so its
Table~\ref{tab:probe-delta} row should be read as one draw from an unstable
training configuration rather than a stable representation property.

\begin{table}[t]
\centering
\caption{Invariance-loss accuracy delta on pose $120^{\circ}$ yaw for
the three rotation-invariant architectures and the two equivariant VN
models (replicate 1), under the trained head and the three
training-free probes. RIConv's, RIConv++'s, and the VN models' pose
accuracy is essentially untouched by the loss. RISurConv pays the
largest pose cost of any architecture in the study, on top of its
representation collapse above.}
\label{tab:supp-probe-pose}
\small
\begin{tabular}{lcccc}
\toprule
Architecture & head$\uparrow$ & sm$\uparrow$ & knn$\uparrow$ & ncm$\uparrow$ \\
\midrule
RIConv & +1.0 & +4.7 & +1.3 & +0.7 \\
RIConv++ & +2.0 & +2.7 & +1.7 & +1.7 \\
RISurConv & $-$13.0 & $-$20.7 & $-$10.3 & $-$12.0 \\
VN-PointNet & +2.7 & $-$0.3 & +0.3 & +2.0 \\
VN-DGCNN & $-$1.3 & $-$5.7 & $-$2.3 & $-$0.7 \\
\bottomrule
\end{tabular}
\end{table}

\paragraph{The RIConv++/PointNet++ crossover is a headroom effect.}
Sec.~\ref{sec:generalization} notes that invariance-trained RIConv++
beats invariance-trained PointNet++ by at least 13 points on noise and
7 on pose under every training-free head, and attributes it to headroom
rather than better generalization. The arithmetic: RIConv++'s untrained
noise accuracy under the trained head is 0.303 against PointNet++'s
0.497, so the identical loss has more room to close on the exact
perturbation it was trained against, and closes enough of it (+52.7
points against PointNet++'s +22.3) to end up ahead on this one matched
condition. This is the same logic as the narrow-transfer law's headroom
condition (Sec.~\ref{sec:generalization}), here playing out across
architectures on a single perturbation instead of across perturbations
on a single architecture. It says nothing about which model
generalizes better to corruptions neither was trained on, which is what
Table~\ref{tab:m40c}'s ModelNet40-C comparison measures instead.

\begin{table*}[t]
\centering
\caption{Head-independent probe (1 of 2, the five standard backbones,
both arms; replicate 1): ModelNet40 held-out accuracy ($n=300$ per
condition) under the trained head, the three training-free probes, and
the Euclidean control. Conditions: clean, noise $\sigma=0.05$, occlusion
camera multiplier 6.0, pose $120^{\circ}$ yaw. Table~\ref{tab:supp-probe2}
continues with the RI family and the equivariant VN models.}
\label{tab:supp-probe}
\footnotesize
\setlength{\tabcolsep}{4pt}
\begin{tabular}{llcccc}
\toprule
Model & Head & Clean$\uparrow$ & Noise .05$\uparrow$ & Occl.\ 6.0$\uparrow$ & Pose 120$\uparrow$ \\
\midrule
PointNet & trained & 0.890 & 0.717 & 0.467 & 0.610 \\
 & subspace & 0.920 & 0.773 & 0.477 & 0.590 \\
 & kNN & 0.927 & 0.763 & 0.490 & 0.593 \\
 & NCM & 0.903 & 0.730 & 0.480 & 0.610 \\
 & NCM (Eucl.) & 0.903 & 0.723 & 0.397 & 0.477 \\
\midrule
PointNet (inv.) & trained & 0.917 & 0.840 & 0.663 & 0.707 \\
 & subspace & 0.927 & 0.827 & 0.640 & 0.697 \\
 & kNN & 0.913 & 0.833 & 0.663 & 0.713 \\
 & NCM & 0.917 & 0.840 & 0.670 & 0.720 \\
 & NCM (Eucl.) & 0.800 & 0.693 & 0.407 & 0.510 \\
\midrule
PointNet++ & trained & 0.883 & 0.497 & 0.290 & 0.770 \\
 & subspace & 0.943 & 0.567 & 0.263 & 0.817 \\
 & kNN & 0.920 & 0.517 & 0.323 & 0.813 \\
 & NCM & 0.897 & 0.507 & 0.317 & 0.780 \\
 & NCM (Eucl.) & 0.890 & 0.463 & 0.390 & 0.767 \\
\midrule
PointNet++ (inv.) & trained & 0.913 & 0.720 & 0.473 & 0.803 \\
 & subspace & 0.950 & 0.583 & 0.513 & 0.870 \\
 & kNN & 0.930 & 0.663 & 0.480 & 0.830 \\
 & NCM & 0.937 & 0.717 & 0.483 & 0.830 \\
 & NCM (Eucl.) & 0.900 & 0.577 & 0.407 & 0.780 \\
\midrule
DGCNN & trained & 0.937 & 0.847 & 0.420 & 0.653 \\
 & subspace & 0.973 & 0.890 & 0.463 & 0.707 \\
 & kNN & 0.957 & 0.843 & 0.477 & 0.663 \\
 & NCM & 0.943 & 0.853 & 0.460 & 0.667 \\
 & NCM (Eucl.) & 0.943 & 0.803 & 0.443 & 0.657 \\
\midrule
DGCNN (inv.) & trained & 0.937 & 0.813 & 0.667 & 0.697 \\
 & subspace & 0.943 & 0.870 & 0.627 & 0.713 \\
 & kNN & 0.957 & 0.870 & 0.657 & 0.707 \\
 & NCM & 0.940 & 0.837 & 0.687 & 0.717 \\
 & NCM (Eucl.) & 0.927 & 0.840 & 0.543 & 0.647 \\
\midrule
PointMLP & trained & 0.927 & 0.580 & 0.393 & 0.867 \\
 & subspace & 0.920 & 0.720 & 0.353 & 0.870 \\
 & kNN & 0.917 & 0.610 & 0.397 & 0.857 \\
 & NCM & 0.927 & 0.583 & 0.397 & 0.853 \\
 & NCM (Eucl.) & 0.930 & 0.537 & 0.337 & 0.860 \\
\midrule
PointMLP (inv.) & trained & 0.927 & 0.850 & 0.743 & 0.903 \\
 & subspace & 0.933 & 0.880 & 0.680 & 0.917 \\
 & kNN & 0.927 & 0.853 & 0.733 & 0.900 \\
 & NCM & 0.927 & 0.860 & 0.753 & 0.903 \\
 & NCM (Eucl.) & 0.840 & 0.657 & 0.483 & 0.783 \\
\midrule
PointTransf. & trained & 0.920 & 0.730 & 0.413 & 0.793 \\
 & subspace & 0.917 & 0.740 & 0.433 & 0.770 \\
 & kNN & 0.917 & 0.740 & 0.413 & 0.790 \\
 & NCM & 0.920 & 0.733 & 0.407 & 0.803 \\
 & NCM (Eucl.) & 0.907 & 0.687 & 0.293 & 0.777 \\
\midrule
PointTransf. (inv.) & trained & 0.933 & 0.850 & 0.657 & 0.833 \\
 & subspace & 0.940 & 0.870 & 0.640 & 0.817 \\
 & kNN & 0.937 & 0.870 & 0.637 & 0.817 \\
 & NCM & 0.933 & 0.853 & 0.653 & 0.833 \\
 & NCM (Eucl.) & 0.903 & 0.833 & 0.400 & 0.757 \\
\bottomrule
\end{tabular}
\end{table*}

\begin{table*}[t]
\centering
\caption{Head-independent probe (2 of 2, the RI family and the
equivariant VN models, both arms; replicate 1): same protocol and
conditions as Table~\ref{tab:supp-probe}.}
\label{tab:supp-probe2}
\footnotesize
\setlength{\tabcolsep}{4pt}
\begin{tabular}{llcccc}
\toprule
Model & Head & Clean$\uparrow$ & Noise .05$\uparrow$ & Occl.\ 6.0$\uparrow$ & Pose 120$\uparrow$ \\
\midrule
RIConv & trained & 0.903 & 0.437 & 0.297 & 0.903 \\
 & subspace & 0.883 & 0.480 & 0.280 & 0.870 \\
 & kNN & 0.913 & 0.413 & 0.310 & 0.913 \\
 & NCM & 0.910 & 0.420 & 0.303 & 0.907 \\
 & NCM (Eucl.) & 0.867 & 0.280 & 0.247 & 0.873 \\
\midrule
RIConv (inv.) & trained & 0.910 & 0.823 & 0.530 & 0.913 \\
 & subspace & 0.917 & 0.823 & 0.430 & 0.917 \\
 & kNN & 0.930 & 0.820 & 0.503 & 0.927 \\
 & NCM & 0.917 & 0.833 & 0.533 & 0.913 \\
 & NCM (Eucl.) & 0.800 & 0.497 & 0.310 & 0.800 \\
\midrule
RIConv++ & trained & 0.920 & 0.303 & 0.313 & 0.923 \\
 & subspace & 0.933 & 0.380 & 0.427 & 0.913 \\
 & kNN & 0.937 & 0.323 & 0.343 & 0.923 \\
 & NCM & 0.923 & 0.320 & 0.327 & 0.927 \\
 & NCM (Eucl.) & 0.903 & 0.307 & 0.247 & 0.917 \\
\midrule
RIConv++ (inv.) & trained & 0.933 & 0.830 & 0.777 & 0.943 \\
 & subspace & 0.940 & 0.830 & 0.747 & 0.940 \\
 & kNN & 0.943 & 0.847 & 0.747 & 0.940 \\
 & NCM & 0.937 & 0.847 & 0.777 & 0.943 \\
 & NCM (Eucl.) & 0.890 & 0.660 & 0.517 & 0.897 \\
\midrule
RISurConv & trained & 0.873 & 0.367 & 0.427 & 0.863 \\
 & subspace & 0.847 & 0.337 & 0.390 & 0.840 \\
 & kNN & 0.850 & 0.380 & 0.413 & 0.843 \\
 & NCM & 0.873 & 0.367 & 0.410 & 0.863 \\
 & NCM (Eucl.) & 0.873 & 0.373 & 0.420 & 0.863 \\
\midrule
RISurConv (inv.) & trained & 0.757 & 0.433 & 0.363 & 0.733 \\
 & subspace & 0.643 & 0.340 & 0.300 & 0.633 \\
 & kNN & 0.763 & 0.383 & 0.387 & 0.740 \\
 & NCM & 0.763 & 0.420 & 0.367 & 0.743 \\
 & NCM (Eucl.) & 0.760 & 0.403 & 0.353 & 0.737 \\
\midrule
VN-PointNet & trained & 0.847 & 0.520 & 0.373 & 0.847 \\
 & subspace & 0.873 & 0.610 & 0.317 & 0.873 \\
 & kNN & 0.870 & 0.563 & 0.367 & 0.870 \\
 & NCM & 0.853 & 0.517 & 0.387 & 0.853 \\
 & NCM (Eucl.) & 0.850 & 0.397 & 0.403 & 0.850 \\
\midrule
VN-PointNet (inv.) & trained & 0.873 & 0.737 & 0.563 & 0.873 \\
 & subspace & 0.870 & 0.730 & 0.440 & 0.870 \\
 & kNN & 0.873 & 0.760 & 0.557 & 0.873 \\
 & NCM & 0.873 & 0.773 & 0.597 & 0.873 \\
 & NCM (Eucl.) & 0.850 & 0.523 & 0.527 & 0.850 \\
\midrule
VN-DGCNN & trained & 0.857 & 0.697 & 0.310 & 0.857 \\
 & subspace & 0.883 & 0.750 & 0.320 & 0.883 \\
 & kNN & 0.877 & 0.727 & 0.327 & 0.877 \\
 & NCM & 0.863 & 0.700 & 0.330 & 0.863 \\
 & NCM (Eucl.) & 0.863 & 0.700 & 0.313 & 0.863 \\
\midrule
VN-DGCNN (inv.) & trained & 0.843 & 0.717 & 0.607 & 0.843 \\
 & subspace & 0.837 & 0.703 & 0.513 & 0.827 \\
 & kNN & 0.853 & 0.730 & 0.583 & 0.853 \\
 & NCM & 0.857 & 0.763 & 0.610 & 0.857 \\
 & NCM (Eucl.) & 0.860 & 0.757 & 0.410 & 0.860 \\
\bottomrule
\end{tabular}
\end{table*}

\section{Artifact Availability, Licensing, and Maintenance}
\label{sec:supp-artifact}

\paragraph{Contents.} The artifact comprises (i) the perturbation and
corruption generators (pose, Gaussian noise, HPR occlusion) with fixed
seeds, so every perturbed evaluation set in this paper is rebuilt
deterministically; (ii) the dataset-preparation scripts that construct
each domain's gallery and query splits from the source datasets;
(iii) the statistical pipeline (pilot, power, confirmatory tests,
ANOVA and Friedman machinery) exactly as run; (iv) the training-free
probe tools of Sec.~\ref{sec:design}; (v) all 193 trained checkpoints (including the
20-model study across three replicate seeds and the five
widened-augmentation retrains); and (vi) the per-cell
raw and summary CSVs behind every table and figure, including the
250-cell confidence-interval and 200-cell pilot-calibration tables
(App.~\ref{sec:supp-ci}, App.~\ref{sec:supp-pilot}) and the
drift--accuracy commensurability data
(App.~\ref{sec:supp-commensurability}).

\paragraph{Licensing.} Code, generators, and the statistical pipeline
are released under the MIT license, and trained checkpoints and result
CSVs under CC~BY~4.0. The artifact redistributes no source-dataset
content: ModelNet40, ShapeNet, ABO, ScanObjectNN, and S3DIS are each
obtained by the user under that dataset's own license or access
agreement (several are gated). The artifact stores only instance
identifiers, split lists, preparation scripts, and generator seeds,
never source geometry. Rebuilding any domain therefore requires the
user's own copy of the source data, which is what the per-domain
preparation scripts consume.

\paragraph{Compute cost.} Table~\ref{tab:supp-cost} reports parameter
counts and profiled FLOPs for all 10 architectures at the benchmark's
input size.

\begin{table}[t]
\centering
\caption{Model cost: parameter count and profiled FLOPs for one
1024-point cloud. FLOPs
are marked $\dagger$ because the profiler misses custom point-cloud
operators for every architecture, so the values are lower bounds and
not comparable across architectures.}
\label{tab:supp-cost}
\small
\setlength{\tabcolsep}{4pt}
\begin{tabular}{lcc}
\toprule
Architecture & Params (M)$\downarrow$ & FLOPs (G)$^\dagger\downarrow$ \\
\midrule
PointNet & 3.40 & 0.45 \\
PointNet++ & 1.50 & 0.86 \\
DGCNN & 1.83 & 2.69 \\
PointMLP & 13.92 & 15.67 \\
PointTransf. & 2.22 & 0.07 \\
RIConv & 0.74 & 0.11 \\
RIConv++ & 0.44 & 0.72 \\
RISurConv & 10.82 & 1.27 \\
VN-PointNet & 1.72 & 1.86 \\
VN-DGCNN & 2.92 & 3.46 \\
\bottomrule
\end{tabular}
\end{table}

\paragraph{Hosting, versioning, and maintenance.} The repository is
public and linked from the project page in Sec.~\ref{sec:design}. The
version evaluated in this paper is the tagged release v1.0.0, archived
on Zenodo (DOI
\href{https://doi.org/10.5281/zenodo.23847450}{10.5281/zenodo.23847450}),
so the paper's numbers remain tied to an immutable artifact even if
the repository evolves. All 193 trained checkpoints ship as assets of
that release. The benchmark suite and pipeline are also packaged as
an installable Python library (\texttt{pip install pcrobust}) whose
model zoo downloads each checkpoint on first use and verifies it
against its recorded SHA256, so the suite can be adopted without
cloning the repository. Corrections and extensions land as tagged point releases with
a changelog. The raw CSVs behind published tables are never
overwritten in place: a corrected table ships alongside its
superseded version. Issues are tracked on the repository, and the
authors commit to maintaining the artifact at least through the
review cycle and the year following publication.

\end{document}